\documentclass[lettersize,journal]{IEEEtran}
\usepackage{lipsum}
\usepackage{times}
\usepackage{algorithm}
\usepackage{algorithmic}
\usepackage{color}
\usepackage{amsmath}
\usepackage{amssymb}
\usepackage{multirow}
\usepackage{makecell}
\usepackage{diagbox}
\usepackage{booktabs}
\usepackage{epsfig}
\usepackage{graphicx}
\usepackage{subfigure}
\usepackage[colorlinks,linkcolor=red]{hyperref}
\usepackage[table ]{ xcolor}
\definecolor{mygray}{gray}{.9}
\definecolor{white}{gray}{1}
\definecolor{mygreen}{RGB}{0,175,0}
\usepackage{url}
\usepackage{tablefootnote}
\usepackage{amssymb}
\usepackage{xspace}
\usepackage{cite}
\usepackage{enumitem}

\makeatletter
\DeclareRobustCommand\onedot{\futurelet\@let@token\@onedot}
\def\@onedot{\ifx\@let@token.\else.\null\fi\xspace}

\def\ie{\emph{i.e}\onedot} 
 
\def\etc{\emph{etc}\onedot}

\makeatother

\begin{document}
\title{Rethinking Sampling Strategies for Unsupervised Person Re-identification}

\author{Xumeng Han$^{\dagger}$, Xuehui Yu$^{\dagger}$, Guorong Li, Jian Zhao, Gang Pan, Qixiang Ye,\\ Jianbin Jiao, and Zhenjun Han$^{\ddagger}$ 
\thanks{$^{\dagger}$\ Equal contribution.}
\thanks{$^{\ddagger}$\ Corresponding author: Zhenjun Han.} 
\thanks{X. Han, X. Yu, G. Li, Q. Ye, J. Jiao and Z. Han are with University of Chinese Academy of Sciences (UCAS), Beijing, 101408 China. E-mail: \{hanxumeng19, yuxuehui17\}@mails.ucas.ac.cn, \{liguorong, qxye, jiaojb, hanzhj\}@ucas.ac.cn.}
\thanks{J. Zhao is with Institute of North Electronic Equipment, Beijing, China, and also with Department of Mathematics and Theories, Peng Cheng Laboratory, Shenzhen, China. 
E-mail: zhaojian90@u.nus.edu.}
\thanks{G. Pan is with Tianjin University, Tianjin, 300350 China. E-mail: pangang@tju.edu.cn.}}

%



\maketitle
 
\begin{abstract}
Unsupervised person re-identification (re-ID) remains a challenging task.
While extensive research has focused on the framework design and loss function, this paper shows that sampling strategy plays an equally important role. 
We analyze the reasons for the performance differences between various sampling strategies under the same framework and loss function.
We suggest that deteriorated over-fitting is an important factor causing poor performance, and enhancing statistical stability can rectify this problem.
Inspired by that, a simple yet effective approach is proposed, termed group sampling, which gathers samples from the same class into groups. The model is thereby trained using normalized group samples, which helps alleviate the negative impact of individual samples. 
Group sampling updates the pipeline of pseudo-label generation by guaranteeing that samples are more efficiently classified into the correct classes. 
It regulates the representation learning process, enhancing statistical stability for feature representation in a progressive fashion.
Extensive experiments on Market-1501, DukeMTMC-reID and MSMT17 show that group sampling achieves performance comparable to state-of-the-art methods and outperforms the current techniques under purely camera-agnostic settings.
Code has been available at  \url{https://github.com/ucas-vg/GroupSampling}.
\end{abstract}


%
\IEEEpeerreviewmaketitle


\section{Introduction} \label{sec:introduction}
%
%
%
%

\IEEEPARstart{U}{nsupervised} person re-identification (re-ID) aims to learn inter-class discriminative representations and intra-class affinities for person identification using unlabeled datasets. 
By reducing the amount of human effort required for data annotation, unsupervised person re-ID has the potential to be extended to practical applications.
Existing deep learning based unsupervised person re-ID research can be generally categorized into unsupervised domain adaptation (UDA) methods \cite{PTGAN,eSPGAN,HHL,SSG,MAR,MEB,CR-GAN,MMT,Ad-cluster,zj1,DG-Net++,tip-uda1,tip-uda2,tip-uda3,tip-uda4,tip-uda5,HCD,JVCT,GCL} and fully unsupervised learning (USL) methods \cite{LOMO,Market,OIM,BUC,DBC,tip-u1,SSL,MMCL,HCT,SpCL,CAP,ICE,DSCE,IICS,MPRD,CycAs,tip-u2}. 
Most recent USL methods~\cite{SpCL,ICE,CAP} estimate pseudo-labels for each sample by mining feature similarities, and then the generated pseudo-labels serve as supervision to train the model.
Currently, USL methods have achieved comparable performance to UDA methods or even to supervised methods without using any identity annotation.
In this paper, we focus on investigating fully unsupervised person re-ID.

The recent SpCL~\cite{SpCL} proposes a self-paced contrastive learning framework that substantially improves the performance of unsupervised person re-ID. We take the vanilla contrastive learning framework in SpCL~\cite{SpCL} (\ie, without the self-paced learning module) as our study baseline, termed \emph{contrastive baseline} in this paper. The conventional person re-ID framework learns inter-class discriminative representations and intra-class affinities by constructing triplets (\ie, anchors, positives and negatives). In distinction, the contrastive baseline builds a memory bank~\cite{memorybank} to store all sample features so that positives and negatives can be obtained directly from the memory bank when optimizing each anchor. It means that for the contrastive baseline, there is no need to sample instances from the same class and different classes in one mini-batch via triplet sampling~\cite{triplet1,triplet} (\ie, $P$ classes and $K$ samples per class are randomly selected). 
Surprisingly, the model fails to obtain the desired results when we replaced the original triplet sampling with the most straightforward random sampling.
Such a phenomenon is not intuitively interpretable since each sample is optimized independently, and there is no interaction between samples in feature extraction, positive and negative selection, and loss calculation. Therefore, intuitively, the sampling strategy should not affect the training results.

\IEEEpubidadjcol

This paper attempts to analyze and explain why different sampling strategies cause this anomaly. We argue that when random sampling is adopted, the randomness and tendency of individual samples may dominate the optimization direction of the entire class, resulting in a shift in the feature representation of the class. 
Specifically, the samples with different ground-truth labels are continuously aggregated, and the class gradually loses the individual identity characteristic and tends to deteriorate feature representation.
Supervision with pseudo-labels formed by such classes will cause the model to lose the capability to distinguish between intra-class similarities and inter-class differences, a phenomenon we call \emph{deteriorated over-fitting}.
Correspondingly, the key to the success of triplet sampling is that it selects a certain number of samples from the same class in each mini-batch. The grouping of same-class samples helps reduce the impact of individual samples on model optimization and highlights the overall trend of the class. This allows each class to maintain its internal statistical stability and prevent it from falling into deteriorated over-fitting. Nevertheless, the success of triplet sampling seems to be coincidental, as it is designed to sample positives and negatives to construct triplets rather than grouping samples of the same class. 
In addition, since $K$ in triplet sampling is fixed, undersampling and oversampling may occur. These two cases require a trade-off to achieve the desired performance.

Therefore, a \emph{group sampling} strategy is designed in this paper with the starting point of grouping samples of the same class.
Group sampling highlights the totality of same-class samples and alleviates the impact of individual samples so that the model is optimized in the direction of consistency with the class-wide trend and facilitates the maintenance of the similarity structure within each class. 
At the same time, group sampling helps to maintain the differentiation between classes, thus preventing classes from continuously aggregating, accumulating noise and falling into deteriorated over-fitting. In this way, the model has the opportunity to extract unique identity similarities by exploiting more subtle differences in the existing similarity structure.
Extensive experiments have been conducted to reflect the phenomenon of deteriorated over-fitting and demonstrate the effectiveness of group sampling in maintaining statistical stability and improving the representation capability of the model.

The contributions of this paper are summarized as follows: 
\begin{itemize}[leftmargin=*]
    \item We provide evidence that sampling strategy plays a vital role in the contrastive baseline for unsupervised person re-ID, and investigate the mechanism of why random sampling causes training collapse and why triplet sampling works.

    \item We highlight the shortcomings involved in triplet sampling, and further propose a novel group sampling strategy for unsupervised person re-ID, which addresses the negative effect of deteriorated over-fitting and enhances statistical stability related to the unsupervised model.
    
    \item Extensive experiments on three large-scale datasets~\cite{Market,Duke,MSMT} show that group sampling substantially improves the performance of the contrastive baseline without additional parameters and computation costs, and yields comparable performance to the state-of-the-arts. Moreover, our method outperforms the current techniques under purely camera-agnostic settings.
\end{itemize}

\section{Related Work}

\subsection{Unsupervised Person re-ID}
In recent years, the person re-ID task has been gradually extended to many more challenging sub-tasks, such as cross-modality person re-ID~\cite{crossmodality1,crossmodality2,reid-survey1,crossmodality3}, person re-ID with label noise~\cite{lablenoise1,lablenoise2}, unsupervised person re-ID~\cite{PTGAN,eSPGAN,HHL,MEB,CR-GAN,MMT,Ad-cluster,JVCT,zj1,DG-Net++,tip-uda1,tip-uda2,tip-uda3,tip-uda4,tip-uda5,SSG,MAR,BUC,DBC,tip-u2,tip-u1,SSL,MMCL,HCT,SpCL,DSCE,GCL,CycAs,CAP,IICS,MPRD,ICE,HCD,LOMO,Market,OIM}, \etc.
Unsupervised person re-ID gets rid of the need for accurate annotations in conventional re-ID frameworks, and aims to learn effective feature representations from unlabeled datasets. Most existing unsupervised person re-ID methods can be generally categorized into two paradigms.
The first is the unsupervised domain adaptation (UDA) methods \cite{PTGAN,eSPGAN,HHL,SSG,MAR,MEB,CR-GAN,MMT,Ad-cluster,JVCT,zj1,DG-Net++,tip-uda1,tip-uda2,tip-uda3,tip-uda4,tip-uda5,GCL,HCD}, which transfer knowledge from the labeled source domain to the unlabeled target domain.
The other is the fully unsupervised learning (USL) methods without any identity annotation \cite{LOMO,Market,OIM,BUC,DBC,tip-u2,tip-u1,SSL,MMCL,HCT,CycAs,SpCL,CAP,DSCE,IICS,MPRD,ICE}.
In this paper, we focus on the USL person re-ID.
Recently, many USL methods train models with generated pseudo-labels as supervision~\cite{HCT,MMCL,SpCL,CAP,ICE,DSCE,IICS}. Pseudo-labels can be obtained by pseudo-label generators or by mining soft label information from feature similarities.
MMT~\cite{MMT} proposed to generate more robust soft labels via mutual mean-teaching.
HCT~\cite{HCT} combined hierarchical clustering with hard-batch triplet loss to improve the quality of pseudo-labels.
MMCL~\cite{MMCL} formulated unsupervised person re-ID as a multi-label classification task to progressively seek true labels.
SpCL~\cite{SpCL} adopted the self-paced contrastive learning strategy to form more reliable clusters.
ICE~\cite{ICE} leveraged inter-instance pairwise similarity to boost the contrastive learning.

In contrast, our method attempts to boost the performance of unsupervised person re-ID on the unlabeled dataset without any source domain data or source domain pre-trained model. The pseudo-label-based strategy is adopted, combining with the training method of contrastive learning. Here, a more reasonable sampling strategy is proposed, which achieves better performance than existing methods.

\subsection{Sampling Strategy} 
Sampling is a basic operation for reducing bias during model learning \cite{sampling1}, and it has been studied for stochastic optimization \cite{stochastic} with the goal of accelerating convergence to the same global loss function. A commonly used way is random sampling \cite{ResNet,random-sampling1}. In the literature, different sampling strategies are proposed to facilitate the learning of various loss functions. For contrastive loss, it is common to select randomly from all possible pairs \cite{contrastive1, contrastive2, random-sampling1}, and sometimes with hard negative mining \cite{contrastive3}. For triplet loss, the semi-hard negative mining method was first utilized in FaceNet \cite{Facenet} and is now widely adopted \cite{semi-hard1, semi-hard2}. For the person re-ID task, triplet sampling \cite{triplet,strongbaseline} is widely used for learning with triplet loss. For each training batch, a certain number of identities are randomly selected, and then several images are sampled from each selected identity. 
This sampling strategy guarantees informative positive and negative mining. In this paper, we study the impact of sampling strategy on unsupervised person re-ID learning, and further propose a new sampling strategy, termed group sampling, to ensure model convergence and to improve performance.

\subsection{Unsupervised Feature Learning}
Deep feature learning aims to learn discriminative feature representations such that visually similar samples are pulled closer and dissimilar samples are pushed farther away~\cite{embeddinglearning}.
Unsupervised feature learning usually focuses on selecting appropriate self-supervised pretext tasks, whose targets are automatically generated without manual labeling. The performance of this invented task is usually not crucial, but the learned intermediate representation is concerned, which is expected to carry good semantic or structural meanings and can be beneficial to practical downstream tasks.
Recently, the state-of-the-art methods on unsupervised feature learning are based on contrastive learning~\cite{memorybank,CPC,CMC,Moco,SimCLR}. They adopt a specific pretext task for instance discrimination by treating every image as a distinct class, which demonstrates better performance than other pretext tasks. 
However, unsupervised person re-ID requires measuring the inter-class affinities, so the strategy for instance discrimination is not suitable.
Similar to DeepCluster~\cite{DeepCluster}, our contrastive baseline utilizes feature clustering and pseudo-label optimization as its pretext task.

\section{How Important Is Sampling?}

This section investigates the importance of sampling strategies for unsupervised person re-ID. In Sec.~\ref{sec:baseline-framework}, we first briefly introduce the contrastive baseline. Then, in Sec.~\ref{sec:description-of-sampling-strategy} and \ref{sec:comparison-of-sampling}, we illustrate how important sampling is through experiments. The experiment settings and implementation details are described in Sec.~\ref{sec:datesets} and \ref{sec:details}.

\subsection{Contrastive Baseline}
\label{sec:baseline-framework}
We adopt the contrastive learning framework proposed in SpCL~\cite{SpCL} (without the self-paced learning module) as our baseline, termed \emph{contrastive baseline}, which combines the self-supervised contrastive learning strategies \cite{memorybank,CMC,Moco,SimCLR} and the pseudo-label-based unsupervised person re-ID methods \cite{SpCL,HCT,MMCL,CAP,ICE,DSCE,IICS}. 
It consists of a CNN-based encoder $f_\theta$ with parameters $\theta$ and a feature memory bank $\mathcal{M}$. The encoder maps the unlabeled dataset $\mathcal{X}=\{x_1,x_2,\dots,x_{n}\}$ to obtain a feature set $\mathcal{V}=\{\boldsymbol{v}_1,\boldsymbol{v}_2,\dots,\boldsymbol{v}_{n}\}$, \emph{i.e.},
\begin{equation}\label{equ:embedding-function}
\mathcal{V}=f_\theta(\mathcal{X}).
\end{equation}The memory bank $\mathcal{M}$ is utilized to store and dynamically update the features of each sample.
Our contrastive baseline uses a pseudo-label generator $\mathcal{G}$ to cluster the features in the memory bank, assigning a pseudo-label to each sample.
Adopting $\mathcal{G}$, $\mathcal{X}$ is divided into a cluster set $\mathcal{C}$ and an outliers set $\mathcal{O}$, where $\mathcal{C}$ contains $N_c$ clusters, \ie,
\begin{equation}\label{equ:clusters-define}
\mathcal{C}=\{C_1,C_2,\dots,C_{N_c}\}.
\end{equation}Samples belonging to the same cluster are assigned the same pseudo-label, and each outlier is assigned a individual pseudo-label.
After being assigned a pseudo-label, the instance is used to construct the unified contrastive loss. The unified contrastive loss function in \cite{SpCL} is defined as,
\begin{equation}\label{equ:contrastive-loss} \small
\mathcal{L}_{\boldsymbol{v}}=-\log\frac{\exp(\left \langle \boldsymbol{v}, \boldsymbol{c}^+ \right \rangle/\tau)}{\sum_{k=1}^{N_c}\exp(\left \langle \boldsymbol{v}, \boldsymbol{c}_k \right \rangle/\tau) + \sum_{k=1}^{N_o}\exp(\left \langle \boldsymbol{v}, \boldsymbol{o}_k \right \rangle/\tau)},
\end{equation}where $\boldsymbol{c}^+$ indicates the positive class prototype corresponding to $\boldsymbol{v}$, the temperature $\tau$ is empirically set as 0.05, $\left \langle \cdot,\cdot \right \rangle$ denotes the inner product between two feature vectors to measure their similarity, $N_c$ is the number of clusters and $N_o$ is the number of outliers.
More specifically, if $\boldsymbol{v}$ is a clustered instance, $\boldsymbol{c}^+ = \boldsymbol{c}_k$ is the centroid of the cluster $C_k$ that $\boldsymbol{v}$ belongs to. If $\boldsymbol{v}$ is an un-clustered outlier, we would have $\boldsymbol{c}^+ = \boldsymbol{o}_k$ as the outlier instance feature corresponding to $\boldsymbol{v}$. When unified contrastive calculating the loss, $\boldsymbol{c}_k$ is calculated from the features belonging to the cluster $C_k$ stored in the memory bank $\mathcal{M}$, and $\boldsymbol{o}_k$ is the feature directly retrieved from $\mathcal{M}$.
The training process includes two steps: (1) using the features in the memory bank to cluster the training set samples, and (2) optimizing the encoder with the unified contrastive loss and dynamically updating the memory bank with encoded features, where the update formula is defined as,
\begin{equation}\label{equ:memory-update}
\mathcal{M}[\boldsymbol{x}] \gets m \cdot \mathcal{M}[\boldsymbol{x}]+(1-m) \cdot \boldsymbol{v},
\end{equation}where $m \in [0,1]$ is the momentum coefficient for updating the sample features in memory bank is empirically set as 0.2. Alg.~\ref{alg:framework} shows the implementation of the contrastive baseline.

\begin{algorithm}[t]
\caption{Contrastive Baseline}
\small
\label{alg:framework}
\begin{algorithmic}[1]
   \REQUIRE Unlabeled training set $\mathcal{X}$, pseudo-label generator $\mathcal{G}$, initialized encoder $f_\theta$, and memory bank $\mathcal{M}$.
   \ENSURE Optimized encoder $f_\theta$.
   \WHILE{\emph{max epochs not reached}}
   \STATE $\mathcal{Y}=\mathcal{G}(\mathcal{M})$
   \STATE $\mathcal{S}=Sampling(\mathcal{X},\mathcal{Y})$
   \FOR{each mini-batch $\mathcal{B}=\{\boldsymbol{x_i},\boldsymbol{y_i}\}$ in $\mathcal{S}$}
   \STATE $\boldsymbol{v_i}=f_\theta(\boldsymbol{x_i})$
   \STATE Compute $\mathcal{L}_{\boldsymbol{v}}$ by Eq.~(\ref{equ:contrastive-loss}) and update the encoder $f_\theta$;
   \STATE Update the memory bank $\mathcal{M}$ with $\boldsymbol{v_i}$ and $m$ by Eq.~(\ref{equ:memory-update});
   \ENDFOR
   \ENDWHILE
\end{algorithmic}
\end{algorithm}

\renewcommand{\thefootnote}{\fnsymbol{footnote}}
\setcounter{footnote}{1}
\setlength{\tabcolsep}{11pt}
\begin{table}[tp]
\renewcommand\arraystretch{1.62}
\begin{center}       
\caption{Performance comparison of sampling strategies.}
\label{tab:sampling}
\begin{tabular}{c||c|c|c|c}
\specialrule{0.1em}{0pt}{0pt}  
	\multirow{2}{*}[0ex]{Sampling Strategy} & \multicolumn{2}{c|}{Market-1501}  & \multicolumn{2}{c}{DukeMTMC-reID}\\ \cline{2-5}
	    & mAP & top-1 & mAP & top-1 \\
	    \specialrule{0.1em}{0pt}{0pt}
        \rowcolor{mygray}
        Random Sampling & 6.1 & 15.1 & 4.2 & 12.3 \\ 
        Triplet Sampling\tablefootnote{Unless otherwise specified, $K$ is set to 4 by default for triplet sampling. 
        } & 48.8 & 70.5 & 44.1 & 64.4 \\
        \rowcolor{mygray}
        RA Sampling & 13.7 & 28.8 & 5.3 & 12.6 \\
        Group Sampling & \textbf{79.2} & \textbf{92.3} & \textbf{69.1} & \textbf{82.7} \\
\specialrule{0.1em}{0pt}{0pt}   
\end{tabular}
\end{center}
\end{table}

\setlength{\tabcolsep}{11pt}
\begin{table}[tp]
\renewcommand\arraystretch{1.53}
\begin{center}       
\caption{Performance of several unsupervised person re-ID frameworks with random sampling.}
\label{tab:random-sampling}
\begin{tabular}{c||c|c|c|c}
\specialrule{0.1em}{0pt}{0pt}  
	\multirow{2}{*}[0ex]{Framework} & \multicolumn{2}{c|}{Market-1501}  & \multicolumn{2}{c}{DukeMTMC-reID}\\ \cline{2-5}
	    & mAP & top-1 & mAP & top-1 \\
	    \specialrule{0.1em}{0pt}{0pt}
        \rowcolor{mygray}
        SpCL \cite{SpCL} & 7.2 & 18.1 & 3.7 & 9.2 \\
        
        CAP \cite{CAP} & 24.4 & 50.2 & 17.2 & 29.6 \\
        \rowcolor{mygray}
        ICE \cite{ICE} & 14.6 & 31.5 & 3.0 & 6.7 \\
        Contrastive Baseline & 6.1 & 15.1 & 4.2 & 12.3 \\ 
\specialrule{0.1em}{0pt}{0pt}   
\end{tabular}
\end{center}
\end{table}

\subsection{Description of Sampling Strategies} \label{sec:description-of-sampling-strategy}
In this section, we describe several sampling strategies in detail. These include the most commonly used random sampling, triplet sampling~\cite{triplet1,triplet,strongbaseline} which is widely used in person re-ID, repeated augmentation (RA)~\cite{ra} which is somewhat similar in form to triplet sampling, and some other sampling strategies~\cite{Facenet,sampling1} in metric learning, where triplet sampling is the default approach in the contrastive baseline.

\subsubsection{Random sampling} Random sampling is the simplest and most commonly used sampling strategy in deep learning. It draws samples randomly from the training set, so the sample composition of each mini-batch is completely random. The sampling list $\mathcal{S}$ can be regarded as the result of randomly shuffling the training set $\mathcal{X}$ with the label set $\mathcal{Y}$, \emph{i.e.},
\begin{equation}\label{equ:random-sampling} 
\mathcal{S} = shuffle(\mathcal{X}, \mathcal{Y}).
\end{equation}

\subsubsection{Triplet sampling} \label{sec:triplet-sampling}
Triplet sampling~\cite{triplet1,triplet} is to sample a certain number of positive and negative samples in each mini-batch. Specifically, it performs data sampling in a $P\times K$ fashion, \ie, $P$ classes and $K$ samples per class are randomly selected.
When the number of samples in a class is greater than $K$, only $K$ of them are randomly selected and the rest will be discarded. Conversely, when the number is less than $K$, some samples will be re-sampled.

In SpCL \cite{SpCL}, the implementation of triplet sampling is slightly different from that in conventional person re-ID \cite{triplet,strongbaseline}.
Specifically, it still selects $K$ samples from each cluster $C_i$. However, each un-clustered outlier is regarded as a particular class, which is not re-sampled but only sampled once. In this paper, we follow this improved triplet sampling.

\subsubsection{Repeated augmentation (RA) sampling}
In RA sampling~\cite{ra}, an image mini-batch $\mathcal{B}$ is formed by sampling $\lceil|\mathcal{B}|/m\rceil$ different images from the dataset, and it is transformed by a set of data increments up to $m$ times to fill the mini-batch.
Therefore, in terms of mini-batch construction, RA sampling is very similar to the $P \times K$ fashion of triplet sampling. The difference is that triplet sampling focuses on sampling different instances within the same class, while RA is repeatedly sampled for each instance.

\subsubsection{Some other sampling strategies}
In deep metric learning, there are many studies devoted to data sampling, where several effective strategies have been designed, such as hard/semi-hard sampling~\cite{Facenet}, distance weighted sampling~\cite{sampling1}, \etc. These sampling strategies focus on how to mine more efficient triplets to facilitate model optimization.
Nevertheless, in implementation, it is difficult to mine triplets directly from all training samples due to the limitation of computational complexity. A common strategy is to sample the data into mini-batches and mine triplets in each mini-batch, where the mini-batch construction follows the $P\times K$ fashion~\cite{triplet1}, \ie, triplet sampling as described above. In contrast, this paper studies data sampling from the perspective of sample composition in each mini-batch. From this perspective, triplet sampling already incorporates the aforementioned sampling strategies in deep metric learning.

\subsection{Comparison of Sampling Strategies} \label{sec:comparison-of-sampling}
The performance obtained using random sampling, triplet sampling, and RA sampling is given in the first three rows of Table~\ref{tab:sampling}. 
It can be clearly seen that triplet sampling yields a great performance gain compared to random sampling.
However, as illustrated in Sec.~\ref{sec:introduction}, the sampling strategy in the contrastive baseline does not directly affect the optimization of each sample.
Therefore, it is not intuitively interpretable why random sampling leads to training collapse and why triplet sampling can somewhat mitigate this phenomenon.
The performance of several unsupervised person re-ID frameworks~\cite{SpCL,ICE,CAP} with random sampling is shown in Table~\ref{tab:random-sampling}.
They yield a significant degradation compared to the reported performance (see Table~\ref{tab:performance}), suggesting that the training collapse caused by random sampling does not only occur in our contrastive baseline but appears to be a relatively common phenomenon.
Moreover, although formally similar to triplet sampling, RA sampling~\cite{ra} fails to yield satisfactory performance, indicating that excessive repeated sampling is not suitable for the contrastive baseline.

\section{Why Triplet Sampling Works?} \label{sec:why-triplet-sampling-works}
In this section, we study why random sampling causes the contrastive baseline to fail to converge and why triplet sampling can alleviate training collapse and guarantee good performance. In Sec.~\ref{sec:deteriorated-overfitting}, we proposed a concept called deteriorated over-fitting and explained the reason for the training collapse caused by random sampling. Then, in Sec.~\ref{sec:statistical-stability}, we propose a corresponding concept called statistical stability and explain that statistical stability is an important factor in restricting deteriorated over-fitting.  Finally, in Sec.~\ref{sec:group-samples-can-achieve}, we explained and analyzed that the grouping operation helps maintain statistical stability, which is why triplet sampling can suppress training collapse.

\subsection{Deteriorated Over-fitting}
\label{sec:deteriorated-overfitting}

\subsubsection{Description of deteriorated over-fitting}
The contrastive baseline utilizes feature similarity and difference between samples to divide samples into different classes, and assigns pseudo-labels to each class. 
However, noise is inevitable in the generated pseudo-labels due to the insufficient expressiveness of the features extracted by the model~\cite{MMT,SpCL,HCT,Ad-cluster}.
Using incorrect pseudo-labels as training supervision will mislead the feature learning~\cite{lablenoise1,lablenoise2}. Therefore, the contrastive baseline is regenerated with new pseudo-labels after a certain number of training iterations~\cite{SpCL}. Under this training strategy, the samples are updated in the feature space in a highly random manner.
Due to the randomness of feature representation, samples will have the opportunity to be merged out of their original class and into another class.

The generated classes can be roughly divided into strong and weak classes.
The strong class is characterized by a relatively large number of samples and their compactness in the feature space, which means that the class has a stable structure.
On the contrary, the sample composition of the weak class is looser, which means that its structure is unstable and easily broken.
Therefore, we believe that samples in the weak class are more likely to separate out and then be swallowed by the strong class. The strong class grows in size when it continues to swallow the weak class or other scattered samples.

When a large number of samples with different ground-truth labels exist in a class, the overall features of the class become generalized and coarse, gradually losing the characterization of individual identities, termed deteriorated feature representation. Fitting to such pseudo-labels with a large amount of noise, the model becomes trapped in learning the deteriorated feature representation, thus losing the capability to distinguish intra-class similarities and inter-class differences. We call this phenomenon \emph{deteriorated over-fitting}.

\subsubsection{Negative effects caused by deteriorated over-fitting}
\begin{figure}[t]
\begin{center}
\includegraphics[width=0.98\linewidth]{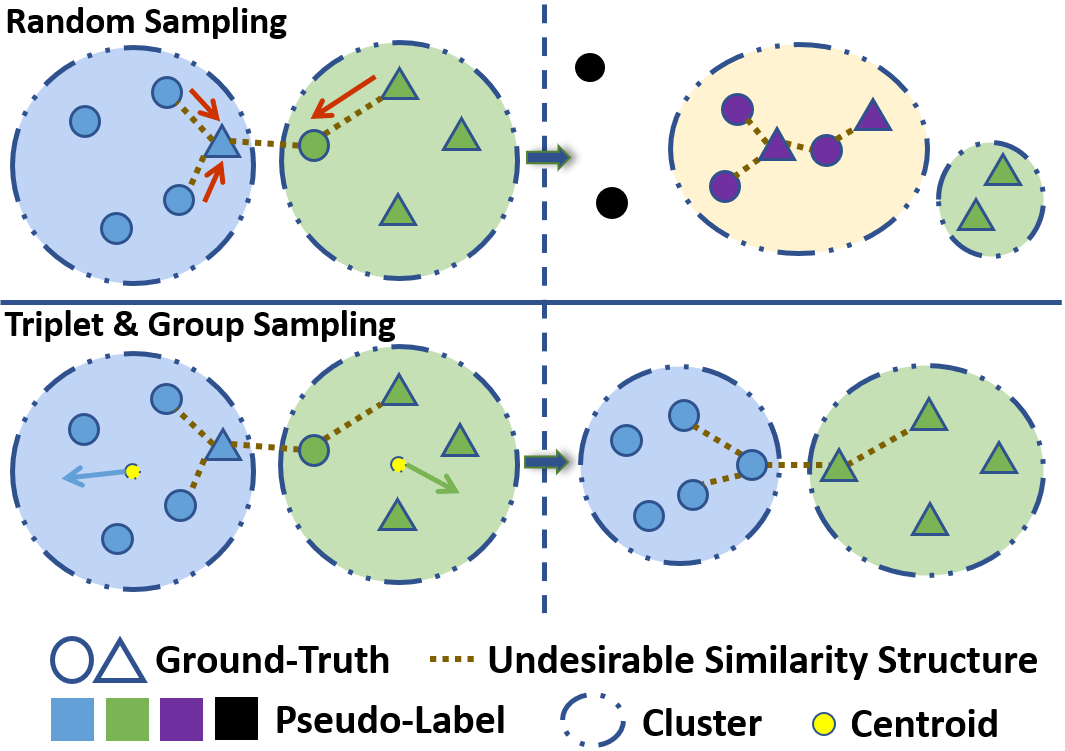}
\end{center}
   \caption{\textbf{(Top)}: Individual samples mislead the optimization trend of overall features of the class, leading to the strengthening of undesirable similarity structure and the destruction of identity structure. As a result, the feature representation tends to deteriorated over-fitting. \textbf{(Bottom)}: Grouping the samples belonging to the same class adopts the overall class trend and weakens the influence of individual samples, thereby forming statistical stability within the class. (Best viewed in color.)} 
\label{fig:motivation}
\end{figure}

The samples in the same class have a certain similarity structure, which encourages them to gather together in the feature space. However, this similarity structure contains not only the unique feature similarity of each person identity, but also many general similarities. We call general similarities undesirable similarities, as shown by the dotted lines in Fig.~\ref{fig:motivation}.

The consequence of deteriorated over-fitting is that various samples with different ground-truth labels are mixed in a class, indicating a continuous aggregation of undesirable similarity structures in the class. In this way, the similarity structure of the identity is submerged in the class and cannot be expressed.
Once such a class is formed, the model learns in the direction of feature deterioration. That is to say, the model can only learn deteriorated semantic information and loses the capability to mine more detailed features between samples. It is particularly disadvantageous for unsupervised person re-ID~\cite{reid-survey1}. Additionally, the knowledge learned from deteriorated classes aggravates the noise in the regenerated pseudo-labels, thus forming a vicious circle.

\subsubsection{Random sampling leads to deteriorated over-fitting}
Randomly selected samples for optimization exacerbates the impact of individual samples on model training. 
It may be manifested as individual samples dominating the optimization direction of the entire class.
Alternatively, some samples break away from the original class, thus disrupting the class structure.
These situations lead to a shift in the feature representation of the class and the reinforcement of undesirable similarity structures. 

For example, as shown in Fig.~\ref{fig:motivation}, the sample features in the class are biased towards undesirable similarity. Once the semantic information represented by this undesirable similarity structure is strengthened, those samples that have been correctly classified but do not have such similarity will be gradually repelled, thereby destroying the similarity structure of identities. Simultaneously, the discrimination between different classes is difficult to maintain, so the strong classes slowly swallow the weak classes, and undesirable similarities within classes continue accumulating.
The ultimate result is the deterioration of semantic information, directly manifested by the increased noise in the pseudo-labels. This is the reason for training collapse caused by random sampling.

\subsubsection{Degree of purity and chaos} \label{sec:purity-chaos}
In order to visually illustrate the phenomenon of deteriorated over-fitting and to prove that random sampling does cause deteriorated over-fitting, we define the degree of purity and chaos.

Intuitively speaking, when most of the samples in a class have the same ground-truth label, it means that the class is relatively pure. By contrast, if there are many samples in a class associated with different ground-truths, the class is comparatively messy. 
We define the degree of chaos $\mathcal{D}_c$ by the average number of identities of all classes, \ie,
\begin{equation}\label{equ:proportion}
\mathcal{C}_i = \mathcal{I}_1^i\cup \mathcal{I}_2^i\cup \dots \cup \mathcal{I}_{n_i}^i,
\end{equation}
\begin{equation}\label{equ:chaos}
\mathcal{D}_c = \frac{1}{N_c} \sum_{i=1}^{N_c} n_i,
\end{equation}where $\mathcal{I}^i_j$ indicates the $j$-th identity set in class $\mathcal{C}_i$, $n_i$ represents the total number of identities in $\mathcal{C}_i$.
We define the degree of purity $\mathcal{D}_p$ by the average proportion of the most numerous identities of all classes, \ie,
\begin{equation}\label{equ:purity}
\mathcal{D}_p= \frac{1}{N_c} \sum_{i=1}^{N_c}\frac{\max\limits_{j} |\mathcal{I}_j^i|}{|\mathcal{C}_i|},
\end{equation}where $|\cdot|$ denotes the number of samples in the set. The degree of purity and chaos reflects the distribution of samples within a class from two perspectives, and they complement each other to reflect the quality of the class. 

\begin{figure}[t]
\begin{center}
\subfigure[Degree of Purity]{
    \centering \hspace{-1.8pt}\includegraphics[width=0.91\linewidth]{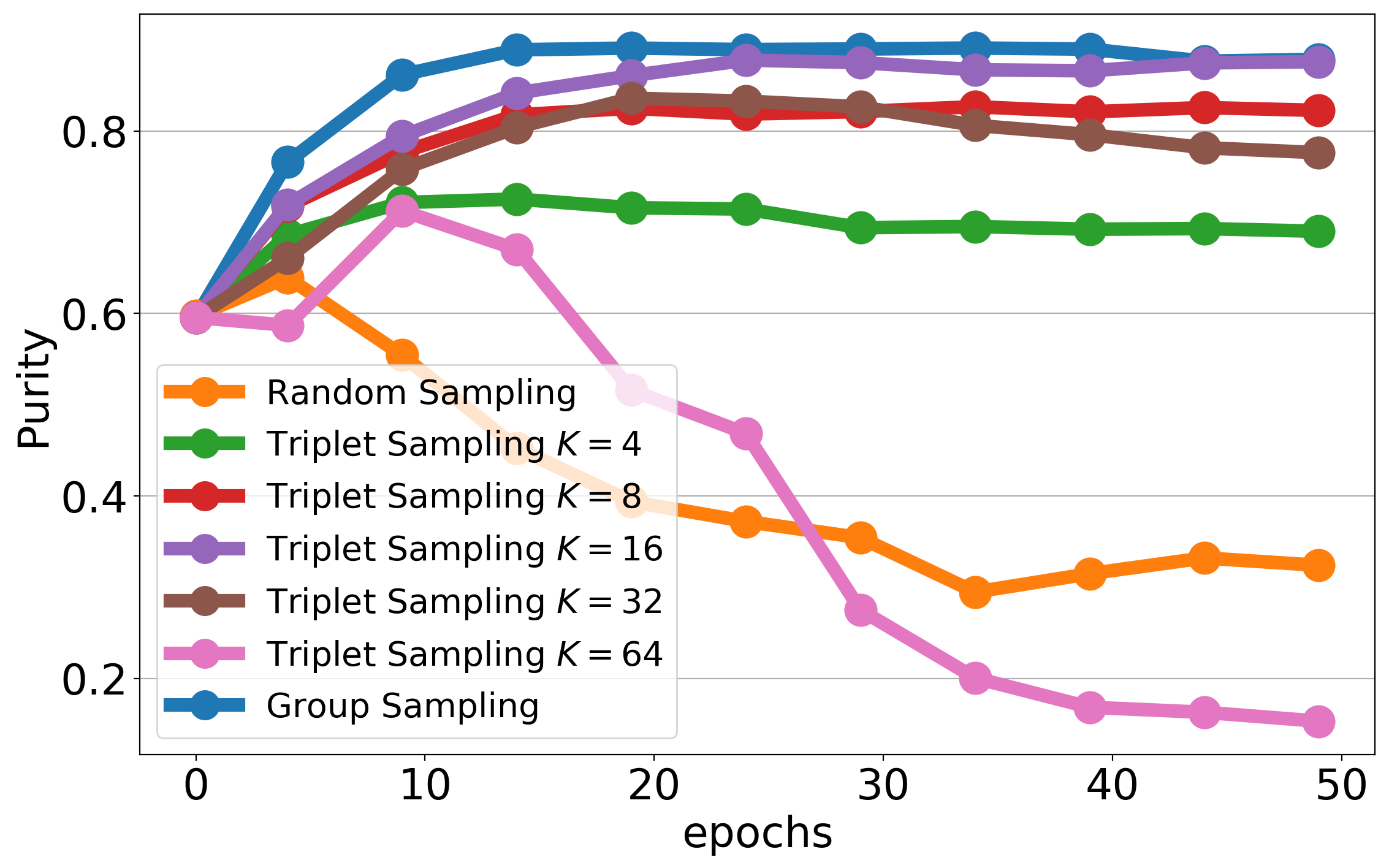}
    \label{fig:purity}
}
\subfigure[Degree of Chaos]{
    \centering
    \includegraphics[width=0.903\linewidth]{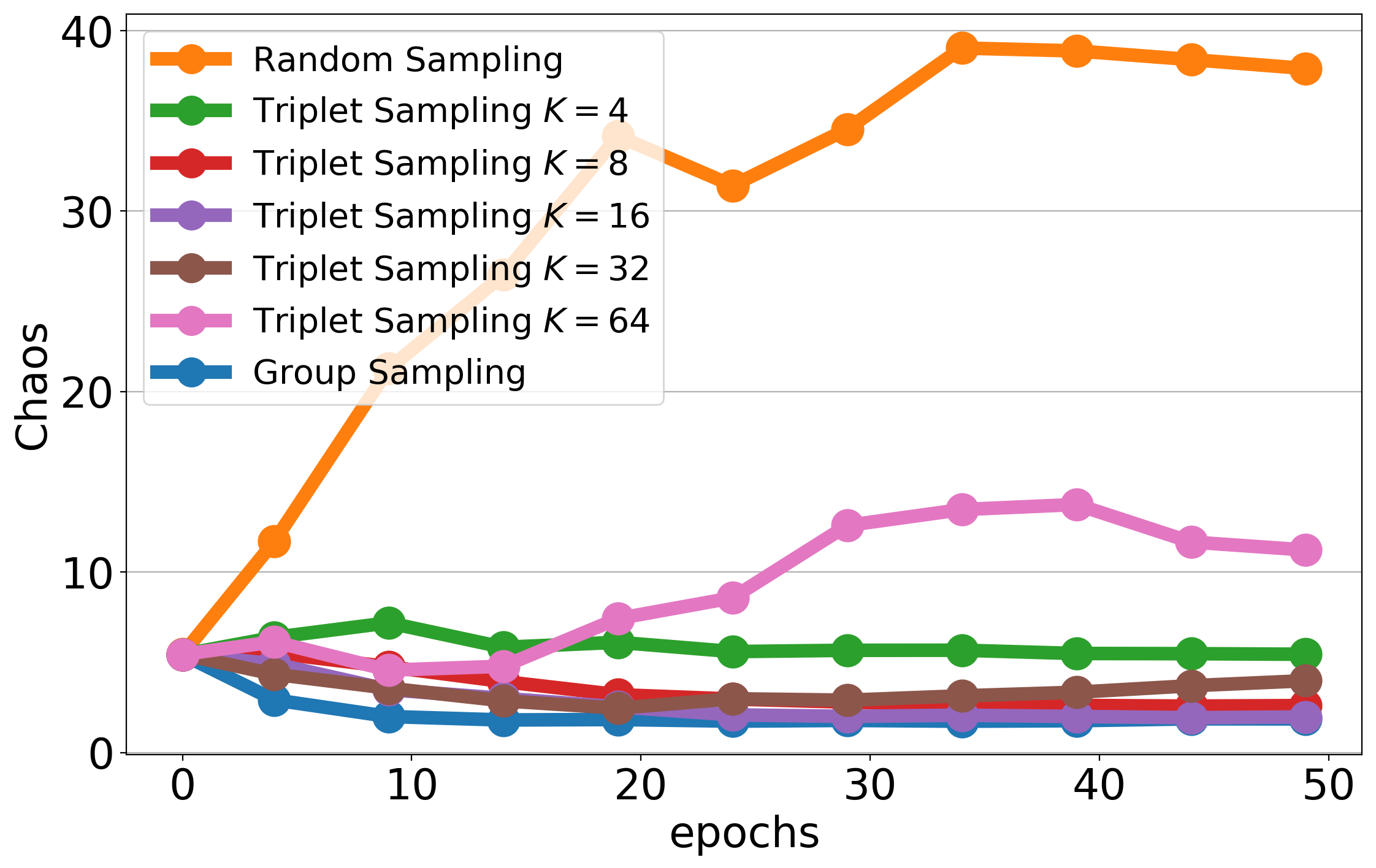}
    \label{fig:chaos}
}
\caption{The degree of purity and chaos for different sampling strategies on Market-1501. It intuitively reflects the deteriorated over-fitting phenomenon caused by random sampling, and triplet sampling and group sampling can suppress this phenomenon. (Best viewed in color.)}
\label{fig:purity-chaos}
\end{center}
\end{figure}

We calculate the degree of purity and chaos of random sampling, as shown in Fig.~\ref{fig:purity-chaos}. It can be seen that random sampling causes many person identities to gather in a class, and no one can dominate the feature representation of this class, which leads to the consequence of deteriorated over-fitting as described in Sec.~\ref{sec:deteriorated-overfitting}.

\subsection{Statistical Stability} \label{sec:statistical-stability}
In the previous section, we explained that the strong classes slowly swallowing the weak classes and undesirable similarities continuously accumulating will lead to deteriorated over-fitting.  
Therefore, to suppress this phenomenon, it is necessary to keep the weak classes stable and less susceptible to breakup during training.
We refer to this process as \emph{statistical stability} maintenance.

The statistical stability guarantees that the similarity structure within the class will not be greatly destroyed, while maintaining a degree of discrimination between other classes. Only if classes maintain their statistical stability can the model continuously explore essential and complex similarities between samples and thus be capable of distinguishing more detailed differences from crude similarities. In this way, the model will learn sufficient inter-class discriminative representations and intra-class affinities to satisfy the need for unsupervised person re-ID. 

\subsection{Grouping Samples Can Achieve Statistical Stability} \label{sec:group-samples-can-achieve}

\subsubsection{Grouping samples is an effective way}
In the previous section, we explained that random sampling leads to deteriorated over-fitting. The main reason is that individual samples affect the overall trend of the class. Therefore, in order to prevent the occurrence of deteriorated over-fitting, it is necessary to reduce randomness and weaken the influence of individual samples. Triplet sampling appears to provide a reasonable solution to this problem.

As described in Sec.~\ref{sec:triplet-sampling}, there is an important parameter $K$ in triplet sampling, \ie, the number of instances per person identity in a mini-batch. This operation enables a group of samples belonging to the same class to be trained at the same time, so the overall trend of the group is emphasized. The randomness of each sample in the group and its own tendency are weakened, and they are not easy to disperse from the group.
The performance gain produced by triplet sampling illustrated in Sec.~\ref{sec:comparison-of-sampling} shows that it can alleviate training collapse, which confirms that the grouping operation in triplet sampling is an effective method to achieve statistical stability.

\subsubsection{Number of instances in triplet sampling affects performance} \label{sec:num-instances}

\setlength{\tabcolsep}{16.5pt}
\begin{table}[tp]
\renewcommand\arraystretch{1.38}
\begin{center}       
\caption{Comparison of different $K$ in triplet sampling.}
\label{tab:num-instances}
\begin{tabular}{c||c|c|c|c}
\specialrule{0.1em}{0pt}{0pt}  
	\multirow{2}{*}[0ex]{$K$} & \multicolumn{2}{c|}{Market-1501}  & \multicolumn{2}{c}{DukeMTMC-reID}\\ \cline{2-5}
	    & mAP & top-1 & mAP & top-1 \\
	    \specialrule{0.1em}{0pt}{0pt}
        \rowcolor{mygray}
        4 & 48.8 & 70.5 & 44.1 & 64.4 \\ 
        8 & 69.4 & 87.2 & 60.5 & 77.5 \\
        \rowcolor{mygray}
        16 & \textbf{77.6} & \textbf{90.2} & \textbf{67.1} & \textbf{81.8} \\
        32 & 46.3 & 71.3 & 42.7 & 62.2 \\
        \rowcolor{mygray}
        64 & 10.2 & 24.4 & 9.6 & 18.1 \\
\specialrule{0.1em}{0pt}{0pt}   
\end{tabular}
\end{center}
\end{table}

We use different values of $K$ to train the contrastive baseline, and the results are shown in Table~\ref{tab:num-instances}.
It can be seen that $K$ is a vital parameter that has a strong impact on the performance. 
When setting $K=16$, a very competitive performance can be obtained. Whereas when $K>16$, there is a significant drop in performance.

\subsubsection{Verification}
In order to visually verify the conclusion that the grouping operation can achieve statistical stability, we calculate the degree of purity and chaos obtained by training with triplet sampling, and also select different values of parameter $K$, as shown in Fig.~\ref{fig:purity-chaos}.

The results show that the qualities of the classes obtained by triplet sampling are consistently better than random sampling. That is, the purity is higher, and chaos is lower. This indicates that the number of person identities in a class is generally small and one (or a few) person identities are dominant in the class. Therefore, the class can better represent the feature of person identity and achieve the desired effect of suppressing deteriorated over-fitting. In addition, it can be seen from the results of different $K$ values that when performance is higher, the degree of purity is higher and chaos is lower. It confirms that a more accurate feature representation of person identity manifests itself when the class suppresses deteriorated over-fitting, resulting in better performance.

\begin{figure*}[t]
\begin{center}
\includegraphics[width=\textwidth]{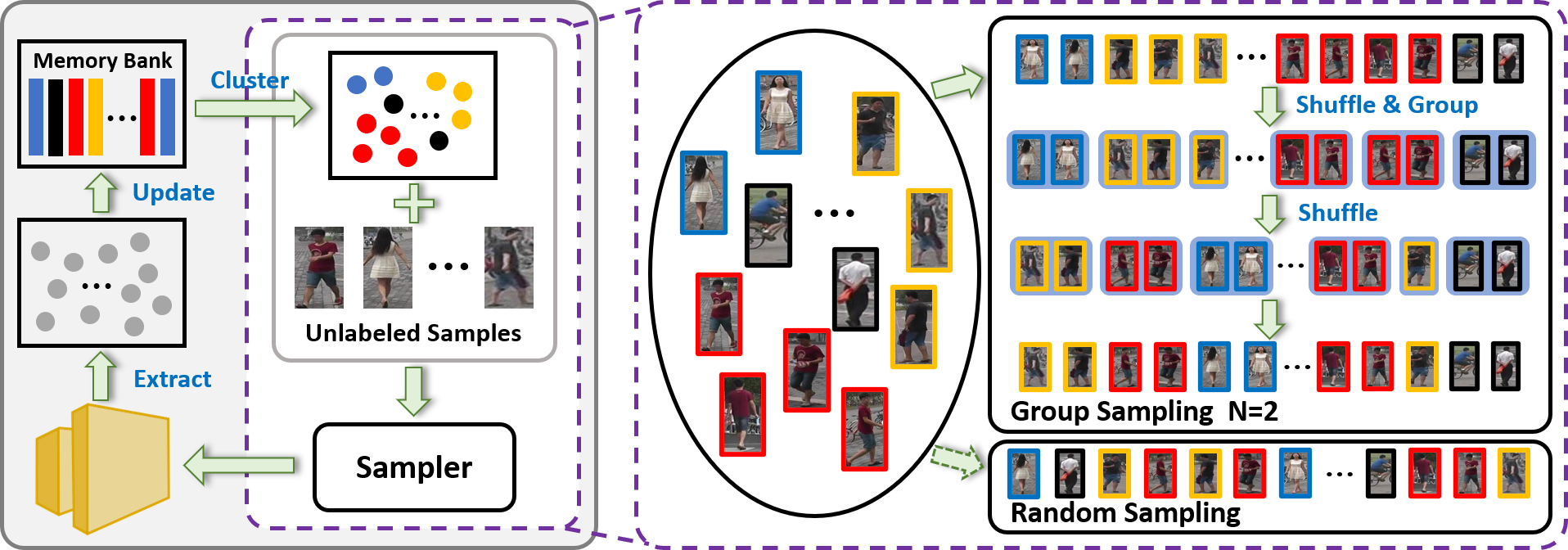}
\end{center}
\caption{\textbf{(Left)}: The framework of the contrastive baseline. The features of each sample are dynamically stored in the memory bank and clustered to generate pseudo-labels. After sampling, the samples are fed into the model for feature extraction and optimized using pseudo-labels. Samples with the same color belong to the same cluster, and the black ones represent outliers.
\textbf{(Right)}: Illustration of group sampling with group size $N=2$, which groups samples belonging to the same class for training. More details are described in Sec.~\ref{sec:group-sampling} and Alg.~\ref{alg:gather-sampling}. (Best viewed in color.)}
\label{fig:framework}
\end{figure*}

\section{What Is A Better Sampling Strategy? }
\label{sec:what-is-a-better-sampling}

\subsection{Problems with Triplet Sampling}
\label{sec:shortcomings}
Despite a significant performance improvement achieved by triplet sampling compared to random sampling on the contrastive baseline, some problems remain.

\subsubsection{Triplet sampling is not designed for grouping}
Although the $P\times K$ fashion of triplet sampling enables the grouping of samples from the same class, this is not its original design intent. The purpose is to sample a sufficient number of positives and negatives (\ie, samples of the same and different classes) in each mini-batch, which is required by the structure of triplet loss. Therefore, the grouping achieved by triplet sampling is an incidental result of sampling positives.

\subsubsection{Triplet sampling requires a trade-off between undersampling and oversampling} \label{sec:shortcomings-1}
As described in Sec.~\ref{sec:description-of-sampling-strategy}, triplet sampling randomly selects a fixed number of samples for each class, which may cause oversampling or undersampling.
When undersampling occurs, some sample features in the memory bank cannot be updated in time. Excessive undersampling is detrimental to training.
When the small number of samples in a class causes oversampling, the sample weights for that class will be amplified, leading to the problem of sample imbalance.
These two cases require a trade-off to achieve the desired performance, as reflected in Table~\ref{tab:num-instances}.
When $K$ is relatively small, undersampling occurs in many classes, thus leading to poor performance.
However, it is not the case that the larger the value, the better the performance. 
When $K$ is larger than the number of samples in many classes, oversampling leads to sample imbalance, which is reflected in the loss calculation and the feature updates in the memory bank. The sample imbalance increases the impact of individual samples, again leading to unsatisfactory performance.

\subsection{Group Sampling}
\label{sec:group-sampling}

\subsubsection{Description}
\label{sec:description-of-group-sampling}

To improve these problems of triplet sampling, we propose a sampling strategy more suitable for the contrastive baseline, termed group sampling. 
We first shuffle the samples in class $C_i$ to increase randomness and then pack each $N$ adjacent samples into a group in each $C_i$, where the hyper-parameter $N$ is the number of samples in each group, named the group size. It is worth emphasizing that when the number of samples in $C_i$ is not divisible by $N$ or less than $N$, the remaining samples are packed directly into a group and not re-sampled.
Finally, all the groups are clustered together and the order of these groups is shuffled. 
Since outliers also significantly affect model training, we treat the outlier set $\mathcal{O}$ as a group and shuffle it with the other groups.
The influence of outliers is further analyzed in Sec.~\ref{sec:influence-outliers} and \ref{sec:treatment-of-outliers}. Alg.~\ref{alg:gather-sampling} describes the implementation details of group sampling, and a schematic diagram is shown in Fig.~\ref{fig:framework} with $N=2$.

\begin{algorithm}[th]
\caption{Group Sampling}
\small
\label{alg:gather-sampling}
\begin{algorithmic}[1]
   \REQUIRE Cluster set $\mathcal{C}$, outlier set $\mathcal{O}$, and group size $N$.
   \ENSURE Sampling sequence $\mathcal{S}$.
   \STATE Shuffle the order of clusters in $\mathcal{C}$;
   \FOR{each cluster $C_i$ in $\mathcal{C}$}
   \STATE Take out all samples in $C_i$ and put them into sample set $\mathcal{X}_i$;
   \STATE Shuffle the order of samples in $\mathcal{X}_i$;
   \WHILE{$|\mathcal{X}_i| > N$}
   \STATE Pack $\{x_1,x_2,\dots,x_N\} \in \mathcal{X}_i$ into a group $G$;
   \STATE Delete $\{x_1,x_2,\dots,x_N\}$ from $\mathcal{X}_i$;
   \STATE Add $G$ to the sampling sequence $\mathcal{S}$;
   \ENDWHILE
   \STATE Group the remaining samples in $\mathcal{X}_i$ and add them to $\mathcal{S}$;
   \ENDFOR
   \STATE Shuffle the order of groups in $\mathcal{S}$;
   \STATE Shuffle the order of outliers in $\mathcal{O}$ and add $\mathcal{O}$ to $\mathcal{S}$;
   \STATE Divide every adjacent $|\mathcal{B}|$ samples in $\mathcal{S}$ into a mini-batch $\mathcal{B}$;
   \STATE Shuffle the order of batches in $\mathcal{S}$.
\end{algorithmic}
\end{algorithm}

\subsubsection{Performance and analysis}
We adopt group sampling to train the contrastive baseline, and the performance is shown in Table~\ref{tab:sampling}. The comparison with random and triplet sampling demonstrates that the grouping operation can prevent training degradation and achieve satisfactory performance.
Furthermore, combining the results in in Table~\ref{tab:num-instances} shows that group sampling also outperforms the optimal performance of triplet sampling, indicating that group sampling can improve the shortcomings of triplet sampling and better match the contrastive baseline.

Similar to Sec.~\ref{sec:purity-chaos}, we plot the purity and chaos curves for group sampling, as shown in Fig.~\ref{fig:purity-chaos}. 
It can be intuitively seen that group sampling yields higher purity and lower chaos compared to random and triplet sampling. 
This indicates that group sampling can achieve the best performance since it prevents the occurrence of deteriorated over-fitting while maintaining the statistical stability within each class. 
In this way, classes gradually learn identity similarity from the initial mixed similarity structure and eventually exhibit identity-unique feature representations.

\begin{figure*}[t]
\begin{center}
\subfigure[mAP score]{
    \centering
    \hspace{-5pt}\includegraphics[width=0.291\textwidth]{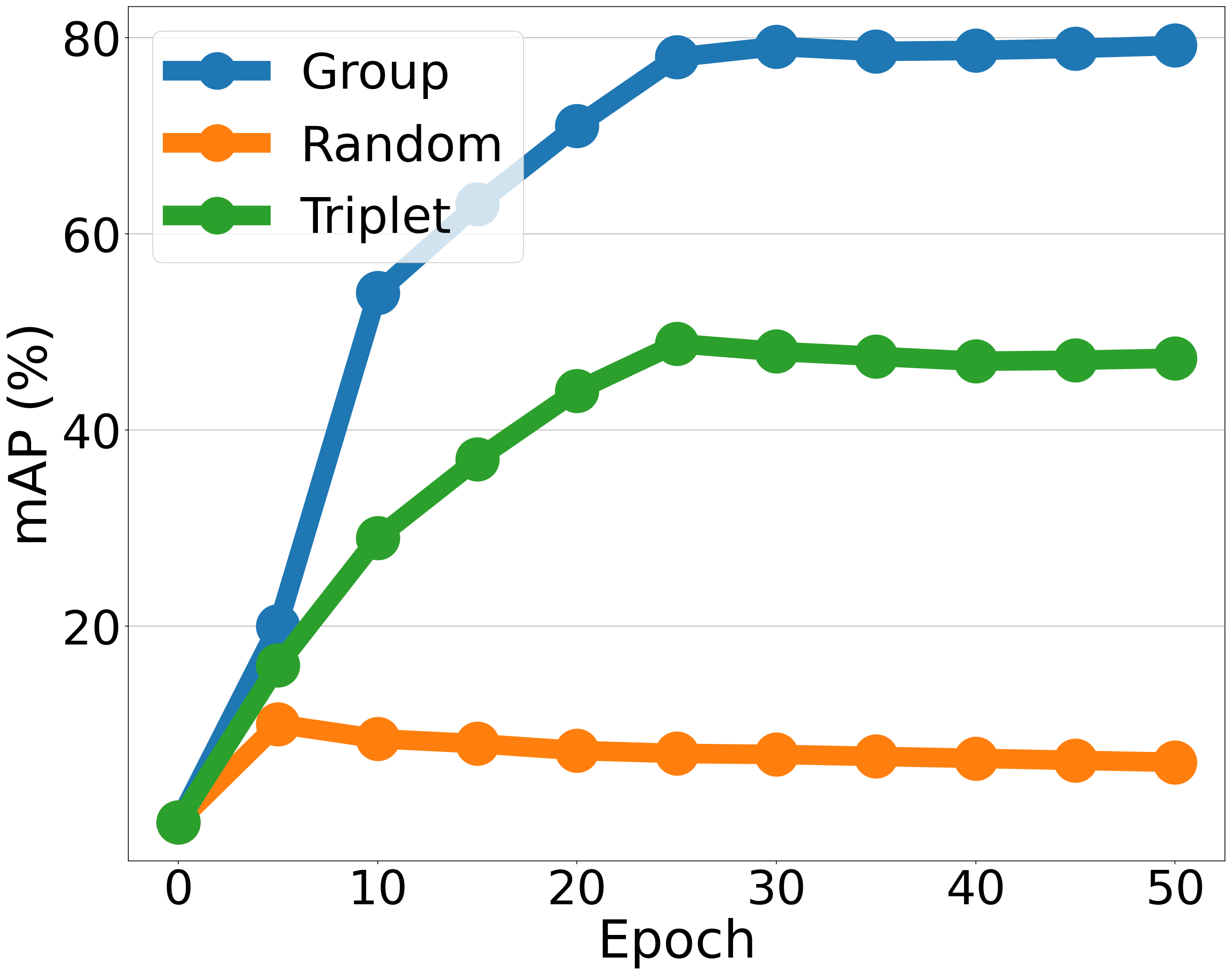}
    \label{fig:map}
}
\subfigure[Number of clusters]{
    \centering
    \includegraphics[width=0.316\textwidth]{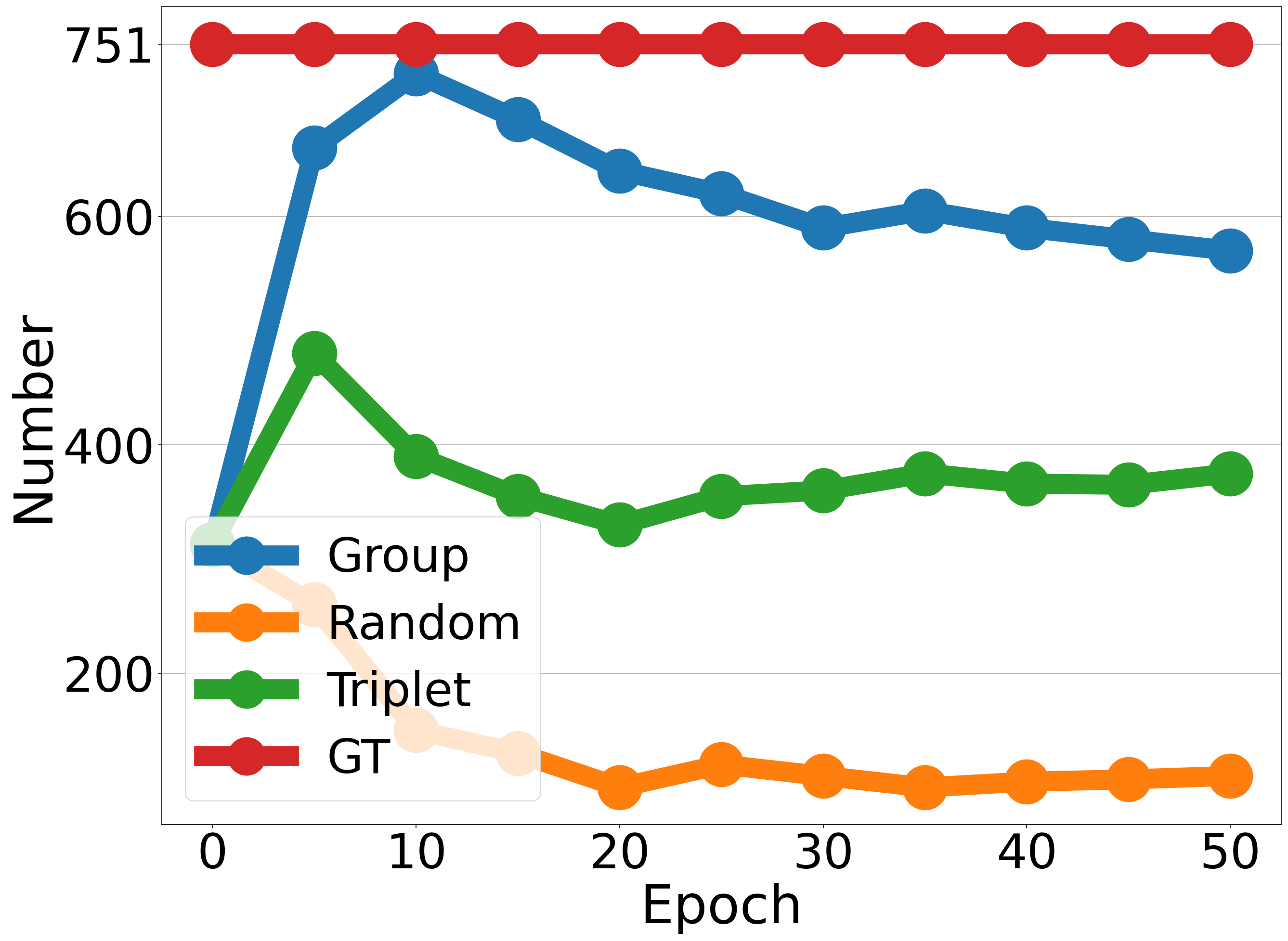}
    \label{fig:number-clusters}
}
\subfigure[NMI score]{
    \centering
    \includegraphics[width=0.313\textwidth]{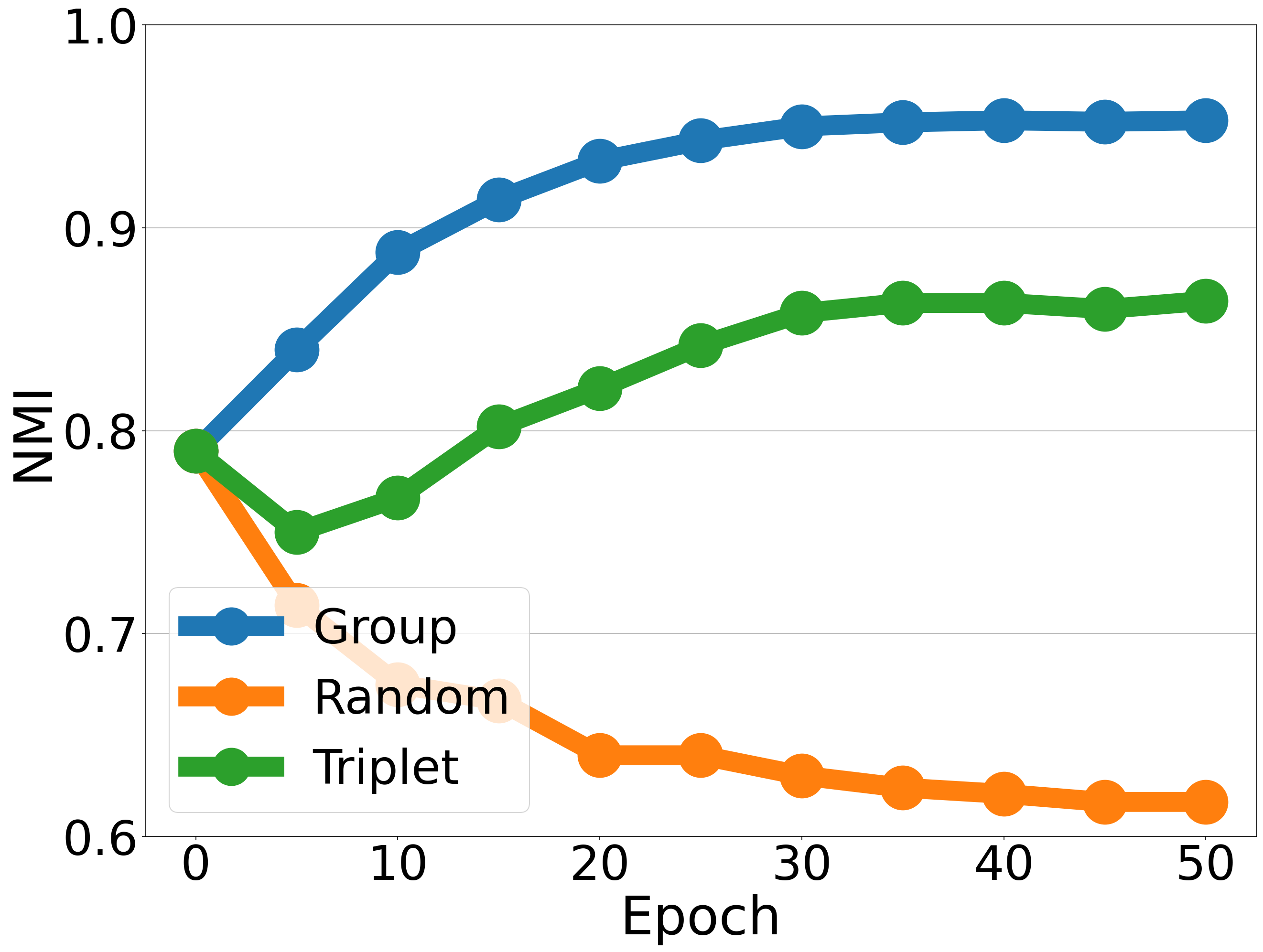}
    \label{fig:nmi}
}
\subfigure[Intra-class and inter-class variance]{
    \centering
    \hspace{4.5pt}\includegraphics[width=0.292\textwidth]{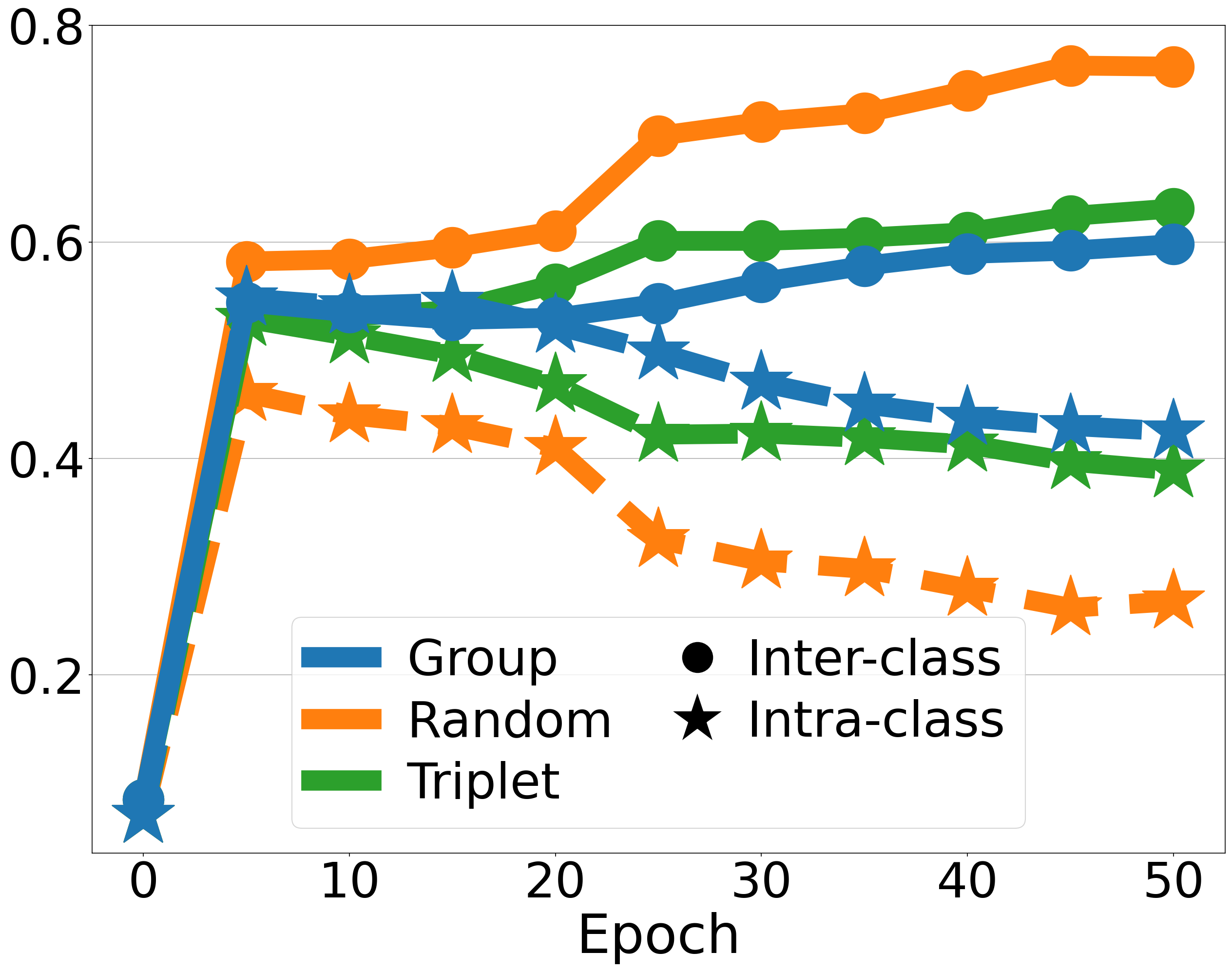}
    \label{fig:variance}
}
\subfigure[Correction rate]{
    \centering
    \includegraphics[width=0.3125\textwidth]{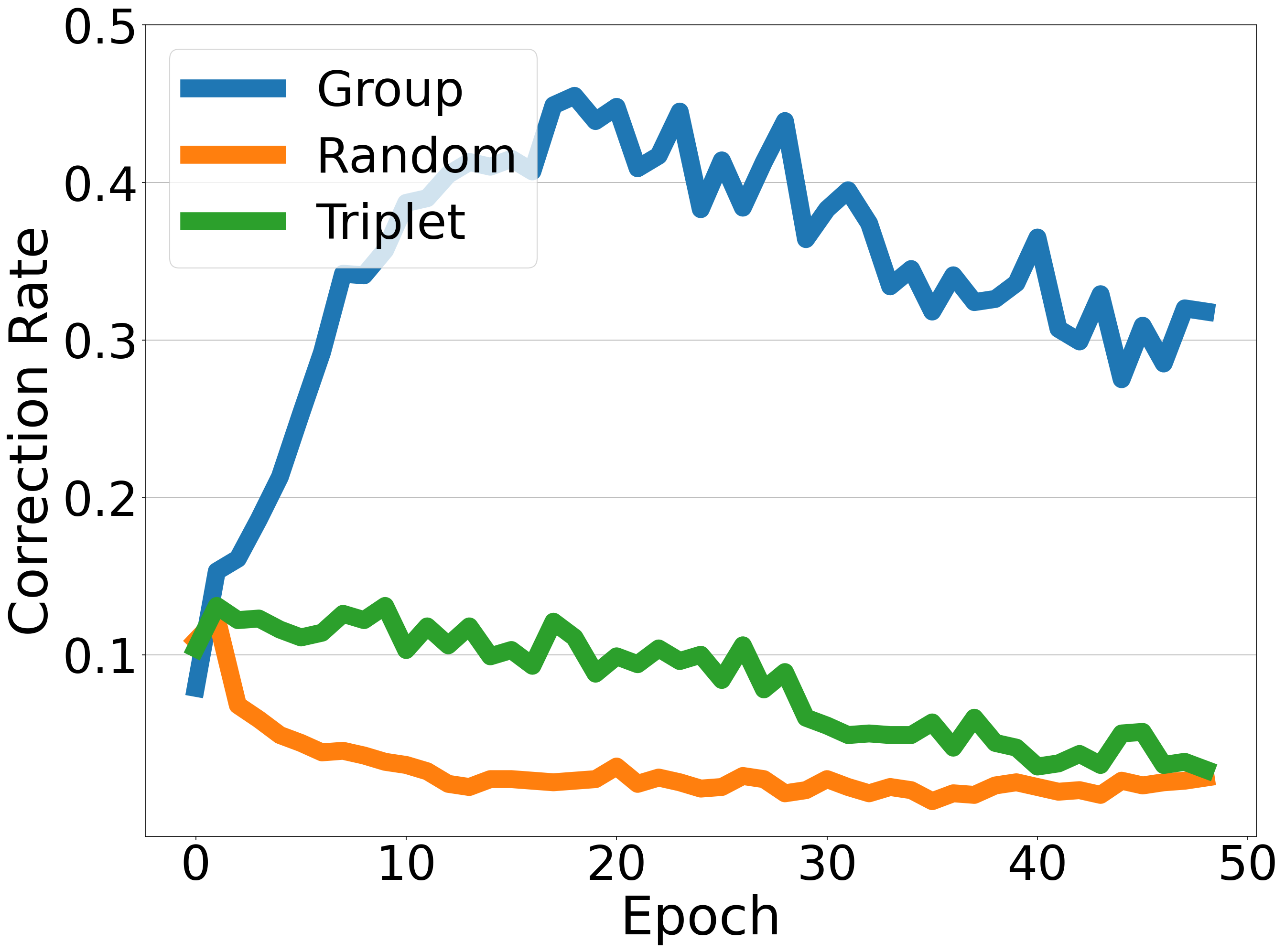}
    \label{fig:correction}
}
\subfigure[Misleading rate]{
    \centering
    \includegraphics[width=0.322\textwidth]{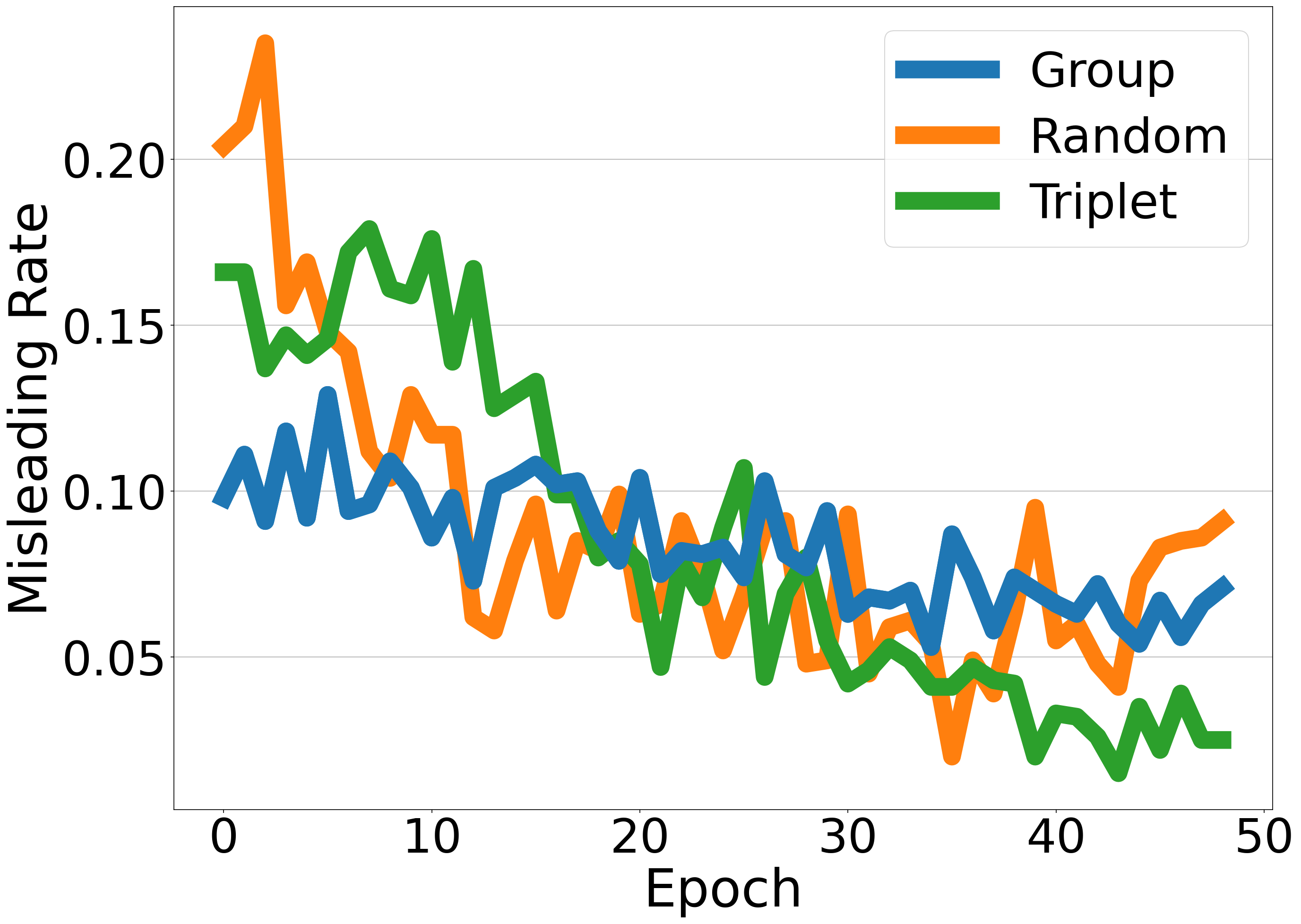}
    \label{fig:misleading}
}
\vspace{-2pt}
\caption{Comparison and analysis between random sampling, triplet sampling and group sampling. It is verified that random sampling leads to low-quality pseudo-labels and the deterioration of class features. In contrast, group sampling maintains the statistically stability within classes, resulting in purer classes and higher quality pseudo-labels. (Best viewed in color.)}
\label{fig:analysis1}
\end{center}
\end{figure*}

\section{Experiment}

\subsection{Datasets and Settings} \label{sec:datesets}

\textbf{Market-1501} \cite{Market} contains 32,668 images of 1,501 identities, each captured by at most 6 cameras. Specifically, it consists of 12,936 training images with 751 identities and 19,732 testing images with 750 identities. All of the images were cropped by a pedestrian detector.

\textbf{DukeMTMC-reID} \cite{Duke} contains 36,411 images with 1,404 identities captured by 8 cameras. Specifically, it contains 16,522 images of 702 identities for training, 2,228 query images and 17,661 gallery images of 702 identities for test.

\textbf{MSMT17} \cite{MSMT} is composed of 126,411 images from 4,101 identities collected by 15 cameras. 
These 15 cameras include 12 outdoor and 3 indoor ones.
Specifically, it contains 32,621 images of 1,041 identities for training, 11,659 query images and 82,621 gallery images for testing.
It suffers from substantial variations of scene and lighting, and is quite challenging. 

\textbf{Evaluation Metrics.} Following \cite{Market}, mean Average Precision (mAP) and Cumulative Matching Characteristic (CMC) are adopted to evaluate the performance. Moreover, results reported in this paper are under the single-query setting, and no post-processing technique is applied.

\begin{figure*}[h]
\begin{center}
\includegraphics[width=0.98\textwidth]{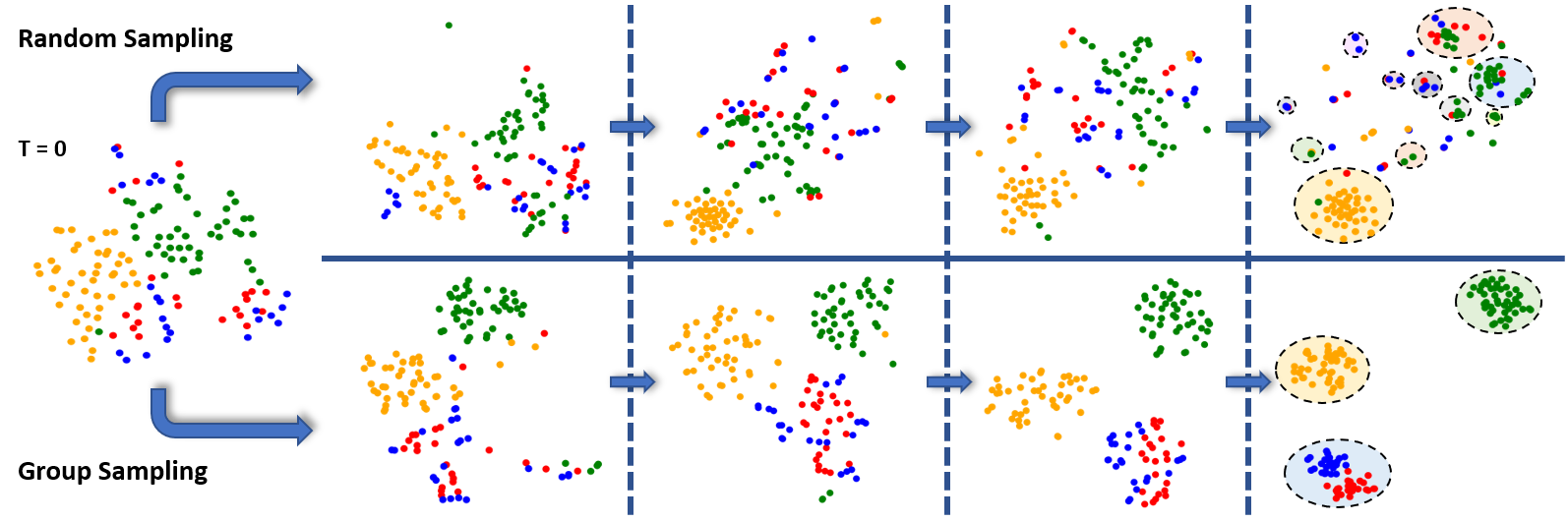}
\vspace{-2pt}
\caption{T-SNE visualization of the distribution of samples in the feature space during training. Samples of the same color belong to the same identity. Dotted circles indicate clusters, and samples not classified into clusters are outliers. It intuitively shows that group sampling successfully gathers samples with the same identity compared to random sampling, indicating that it facilitates enhanced feature representation. (Best viewed in color.)}
\label{fig:tsne}
\end{center}
\end{figure*}

\subsection{Implementation Details} \label{sec:details}

We adopt ResNet-50~\cite{ResNet} pre-trained on ImageNet~\cite{ImageNet} as the backbone of the encoder $f_\theta$. The subsequent layers after pooling-5 layer are removed, and a 1D batch normalization layer and an $\mathcal{L}_2$-normalization layer are added. 
DBSCAN \cite{DBSCAN} is used as the pseudo-label generator $\mathcal{G}$ and Jaccard distance with $k$-reciprocal nearest neighbors \cite{rerank} is used for clustering before each epoch, where we set $k=30$. For DBSCAN, the maximum distance between neighbors is set as $d=0.6$ and the minimal number of neighbors for a dense point is set as 4. The temperature $\tau$ in unified contrastive loss is set as 0.05, and the momentum coefficient in memory bank is set as 0.2. The number of training epochs is set to 50, the initial learning rate is set to 0.00035 and is divided by 10 after every 20 epochs. The batch size is set to 64. The input images are resized to 256$\times$128. Strategies like randomly flipping, cropping and erasing \cite{erasing} are also introduced.
For Market-1501, DukeMTMC-reID and MSMT17, the group size $N$ is set to 256, 128 and 1,024, respectively. More ablation studies on group size $N$ are discussed in Sec.~\ref{sec:param-N}.

For Market-1501 and DukeMTMC-reID, it takes approximately 2 hours for training with a single GTX 2080TI GPU. 
For MSMT17, we use 4 GPUs for training due to its large data volume, and the total training time is about 5 hours. Further, we adjusted some hyper-parameters for it to achieve better performance. Specifically, the initial learning rate is set to 0.00005, the batch size is set to 256, and the momentum coefficient in the memory bank is set to 0.1.

\subsection{Analysis} \label{sec:analysis}
In Sec.~\ref{sec:why-triplet-sampling-works} and \ref{sec:what-is-a-better-sampling}, we describe the deteriorated feature representation caused by random sampling, the statistical stability of triplet sampling and group sampling, and the shortcomings in triplet sampling.
In this section, we further analyze and compare the performance of the three sampling strategies in terms of mAP score, number of clusters and Normalized Mutual Information (NMI) score, intra-class and inter-class variance, correction rate, and misleading rate, as shown in Fig.~\ref{fig:analysis1}.
All experiments are conducted on Market-1501 with unsupervised person re-ID settings.

\subsubsection{Number of clusters and NMI score} \label{sec:number-nmi}
The number of clusters generated by clustering and NMI score can intuitively reflect the quality of pseudo-labels, as shown in Fig.~\ref{fig:number-clusters} and~\ref{fig:nmi}. It can be seen that random sampling results in a sharp decrease in the number of clusters and NMI score. Fig.~\ref{fig:number-clusters} illustrates that random sampling results in fewer clusters, indicating that the samples are grouped in large-scale clusters in which the components are more mixed. The NMI score further shows that the reliability of the pseudo-labels generated by random sampling is very low.
In contrast, triplet sampling and group sampling produce clusters that are closer to the ground-truth number of identities and obtain higher NMI scores. Our group sampling yields the highest NMI score with a maximum of 0.95, which indicates more reliable clustering and higher quality pseudo-labeling results, and also intuitively reflects the reason why group sampling yields the best performance.

\subsubsection{Intra-class and inter-class variance}
The intra-class variance of samples within a class and the inter-class variance between classes are calculated to quantitatively analyze the degree of aggregation and dispersion of samples in the feature space. Fig.~\ref{fig:variance} shows the curves for intra-class and inter-class variance during training. It can be seen that as the training progresses, the intra-class variance is constantly decreasing, and the inter-class variance is increasing. It shows that samples in the feature space are gradually grouped into compact clusters from a relatively dispersed state at the beginning, and the distance between clusters is constantly increasing.

However, these three sampling strategies produced slightly different variance curves. Random sampling results in tighter samples in each cluster and more dispersion of different clusters, which leads to samples with the same ground-truth being scattered and thus losing the opportunity to be corrected once they are misclassified. In contrast, triplet sampling and group sampling the clusters more internally loose and closer to each other, so as to achieve relatively soft margin. Samples assigned with incorrect pseudo-labels have more chances to break away from the wrong class and be classified into the correct class. In this way, the reliability of pseudo-labels is improved, and then the representational capability of the model is iteratively improved. This corroborates our elaboration that the grouping operation is an effective way to suppress deterioration.

\subsubsection{Correction rate and misleading rate}
In this part, we define the correction rate and the misleading rate to further analyze the difference between the three sampling strategies. We find the identity set $\mathcal{I}_{j^*}^i$ with the largest proportion in each class $\mathcal{C}_i$, as expressed in Eq.~(\ref{equ:purity}), where $j^*$ represents the principal identity of this class. We subjectively define a sample to be correctly classified if the identity $j$ of the sample is consistent with the principal identity $j^*$ of the class.
We compare the classification results of samples in two adjacent epochs. If the incorrectly classified sample in the previous epoch is classified correctly in this epoch, we call it correction. On the contrary, if the correctly classified sample in the previous epoch is incorrectly classified in this epoch, we call it misleading. 

The curves of correction rate and misleading rate are shown in Fig.~\ref{fig:correction} and \ref{fig:misleading}.
It illustrates that random sampling causes the correction rate remains low during the training process, which makes the initial misclassification of samples almost impossible to rectify. At the same time, in the early stage of training, the misleading rate is relatively high, which causes an increasing number of errors to accumulate. This makes the overall semantic information of the class to deviate and leads to deteriorated over-fitting. This phenomenon confirms the description in Sec.~\ref{sec:deteriorated-overfitting}. The decrease in the misleading rate in the latter stage of training is because the overall structure of a class has been formed, and there are few sample interactions between classes.
In contrast, triplet sampling has a slightly higher correction rate than random sampling, while group sampling has the highest. The high correction rate indicates that the initially misclassified samples have more chances to be corrected, which helps to gradually improve the quality of pseudo-labels. 
Meanwhile, the misleading rate of group sampling has been in a low range, which inhibits the degradation of semantic information of the classes.

\setlength{\tabcolsep}{15.9pt}
\begin{table}[t]
\renewcommand\arraystretch{1.3108}
\begin{center}
\caption{Ablation study on group size $N$ in group sampling.}
\label{tab:parameter}
\begin{tabular}{c||c|c|c|c}
\specialrule{0.1em}{0pt}{0pt}
\multirow{2}{*}[0ex]{$N$} & \multicolumn{2}{c|}{Market-1501} & \multicolumn{2}{c}{DukeMTMC-reID} \\ \cline{2-5}
 & mAP & top-1 & mAP & top-1 \\
\specialrule{0.1em}{0pt}{0pt}  
\rowcolor{mygray} 1 & 6.1 & 15.1 & 4.2 & 12.3 \\
2 & 13.1 & 28.2 & 5.6 & 12.2 \\
\rowcolor{mygray} 4 & 20.2 & 37.4 & 8.0 & 16.7 \\
8 & 44.6 & 66.4 & 15.1 & 26.0 \\
\rowcolor{mygray} 16 & 64.5 & 82.9 & 54.3 & 72.0 \\
32 & 76.7 & 90.2 & 66.8 & 81.6 \\
\rowcolor{mygray} 64 & 78.1 & 90.7 & 68.7 & \textbf{83.1} \\
128 & 78.5 & 91.5 & \textbf{69.1} & 82.7 \\
\rowcolor{mygray} 256 & \textbf{79.2} & \textbf{92.3} & 68.9 & 82.0 \\
512 & 78.9 & 91.9 & 68.4 & 82.0 \\
\rowcolor{mygray} 1,024 & 79.1 & 91.6 & 69.1 & 82.5 \\
\specialrule{0.1em}{0pt}{0pt}  
\end{tabular}
\end{center}
\end{table}

\setlength{\tabcolsep}{15.2pt}
\begin{table}[t]
\renewcommand\arraystretch{1.33}
\begin{center}
\caption{Ablation study on shuffling degree $M$ on Market-1501.}
\label{tab:degree-of-sampler-shuffle}
\begin{tabular}{c||c|c|c|c}
\specialrule{0.1em}{0pt}{0pt}  
$M$ & mAP & top-1 & \# clusters & NMI \\ 
\specialrule{0.1em}{0pt}{0pt}   
\rowcolor{mygray}
1  & \textbf{79.2} & \textbf{92.3} & \textbf{607} & \textbf{0.95} \\  
4   & 65.0 & 84.1 & 436 & 0.89 \\ 
\rowcolor{mygray}
16  & 16.6 & 35.5 & 210 & 0.71 \\ 
64  & 5.9 & 15.8 & 106 & 0.66 \\
\rowcolor{mygray}
$all$  & 6.1 & 15.1 & 104 & 0.62 \\ 
\specialrule{0.1em}{0pt}{0pt}   
\end{tabular}
\end{center}
\end{table}

\setlength{\tabcolsep}{6pt}
\begin{table}[t]
\renewcommand\arraystretch{1.5}
\begin{center}
\caption{Ablation study on influence of outliers. Clusters: Only clustered instances are used for training. Clusters+Outliers: Including un-clustered outliers into training data.}
\label{tab:outliers}
\begin{tabular}{c|c||c|c|c|c}
\specialrule{0.1em}{0pt}{0pt}
\multirow{2}{*}[0ex]{Sampling} & \multirow{2}{*}[0ex]{Method} & \multicolumn{2}{c|}{Market-1501} & \multicolumn{2}{c}{DukeMTMC-reID} \\ 
\cline{3-6}
 & & mAP & top-1 & mAP & top-1 \\
\specialrule{0.1em}{0pt}{0pt}
\rowcolor{mygray}
\cellcolor{white}
\multirow{2}{*}[0ex]{Triplet} & Clusters & 3.2 & 8.6 & 2.1 & 5.8 \\
& Clusters\,+\,Outliers & 48.8 & 70.5 & 44.1 & 64.4 \\
\specialrule{0.1em}{0pt}{0pt}
\rowcolor{mygray}
\cellcolor{white}
\multirow{2}{*}[0ex]{Group} & Clusters & 9.7 & 23.2 & 4.1 & 9.6 \\
 & Clusters\,+\,Outliers & \textbf{79.2} & \textbf{92.3} & \textbf{69.1} & \textbf{82.7} \\
\specialrule{0.1em}{0pt}{0pt}
\end{tabular}
\end{center}
\end{table}

\setlength{\tabcolsep}{7.2pt}
\begin{table}[t]
\renewcommand\arraystretch{1.45}
\begin{center}
\caption{Ablation study on outlier treatment in sampling. Tre.~\uppercase\expandafter{\romannumeral1}: Each outlier is treated as an independent group. Tre.~\uppercase\expandafter{\romannumeral2}: The outlier set including all the outliers is treated as a group.}
\label{tab:treatment}
\begin{tabular}{c|c||c|c|c|c}
\specialrule{0.1em}{0pt}{0pt}
\multirow{2}{*}[0ex]{Outliers} & \multirow{2}{*}[0ex]{Sampling} & \multicolumn{2}{c|}{Market-1501} & \multicolumn{2}{c}{DukeMTMC-reID} \\ 
\cline{3-6}
 & & mAP & top-1 & mAP & top-1 \\
\specialrule{0.1em}{0pt}{0pt}
\rowcolor{mygray}
\cellcolor{white}
\multirow{3}{*}[0ex]{Tre.~\uppercase\expandafter{\romannumeral1}} & Triplet $K=4$ & 48.8 & 70.5 & 44.1 & 64.4 \\
 & Triplet $K=16$ & 77.6 & 90.2 & 67.1 & 81.8\\
\rowcolor{mygray}
\cellcolor{white} & Group & 64.2 & 81.1 & 62.8 & 79.3 \\
\specialrule{0.1em}{0pt}{0pt}
\multirow{3}{*}[0ex]{Tre.~\uppercase\expandafter{\romannumeral2}} & Triplet $K=4$ & 52.9 & 75.3 & 46.7 & 65.5 \\
& \cellcolor{mygray}Triplet $K=16$ & \cellcolor{mygray}76.0 & \cellcolor{mygray}90.2 & \cellcolor{mygray}66.9 & \cellcolor{mygray}81.3 \\
 & Group & \textbf{79.2} & \textbf{92.3} & \textbf{69.1} & \textbf{82.7} \\ 
\specialrule{0.1em}{0pt}{0pt}
\end{tabular}
\end{center}
\end{table}

\subsection{Visualization Analysis}
We randomly select four identities and visualize the distribution of the samples in the feature space during training, as shown in Fig.~\ref{fig:tsne}. For random sampling, the samples of these four identities are divided into many clusters, and each cluster contains multiple identity samples. It is worth noting that the number of clusters produced by random sampling is more than that observed in group sampling, which seems to be contrary to statistical results in Sec.~\ref{sec:number-nmi} which shows that random sampling results in fewer clusters. This is because the samples of each identity are scattered in many classes, as illustrated in Fig.~\ref{fig:purity-chaos}, and there are many samples of other identities in each cluster which are not drawn in the figure.

In contrast, it is obvious that group sampling can distinguish samples with different identities, and samples with the same identity can be correctly clustered. Although some samples with different identities (red and blue points) are classified into the same cluster, they are not completely mixed, indicating that their features are still clearly distinguishable. 
It is also important to note that all three clusters contain only the samples shown in the figure and no samples of other identities.

Visualization results further validate that group sampling can prevent samples with the same ground-truth identity from being misclassified into different classes, while avoiding deteriorated over-fitting. Simultaneously, during the training process, the identity similarity structure is not destroyed, which helps the model gradually learn the characteristics of each identity, thereby verifying the importance of statistical stability described in Sec.~\ref{sec:statistical-stability}.

\setlength{\tabcolsep}{7.5pt}
\begin{table*}
\renewcommand\arraystretch{1.5}
\begin{center}
\caption{Comparison of the state-of-the-art unsupervised person re-ID methods on Market-1501, DukeMTMC-reID and MSMT17. The \textcolor{red}{\textbf{first}}, \textcolor{blue}{\textbf{second}} and \textcolor{mygreen}{\textbf{third}} best results are marked in red, blue and green, respectively. The best performance under purely camera-agnostic settings is marked in \textbf{\underline{underline}}. $\dagger$ denotes the camera-aware methods. $\ast$ denotes the performance under purely camera-agnostic settings implemented according to the authors' code, as detailed in Sec.~\ref{sec:sota}.}
\label{tab:performance}
\begin{tabular}{l|c||c|c|c|c|c|c|c|c|c|c|c|c}
\specialrule{0.1em}{0pt}{0pt}  
    \multicolumn{2}{c||}{\multirow{2}{*}[0ex]{Method}} & \multicolumn{4}{c|}{Market-1501} & \multicolumn{4}{c|}{DukeMTMC-reID} & \multicolumn{4}{c}{MSMT17}  \\ \cline{3-14}
\multicolumn{2}{c||}{} & mAP & top-1 & top-5 & top-10 & mAP & top-1 & top-5 & top-10 & mAP & top-1 & top-5 & top-10 \\
\specialrule{0.1em}{0pt}{0pt}
\rowcolor{mygray}
LOMO \cite{LOMO} & CVPR'15 & 8.0 & 27.2 & 41.6 & 49.1 & 4.8 & 12.3 & 21.3 & 26.6 & - & - & - & - \\
BOW \cite{Market} & ICCV'15 & 14.8 & 35.8 & 52.4 & 60.3 & 8.3 & 17.1 & 28.8 & 34.9 & - & - & - & - \\
\rowcolor{mygray}
OIM \cite{OIM} & CVPR'17 & 14.0 & 38.0 & 58.0 & 66.3 & 11.3 & 24.5 & 38.8 & 46.0 & - & - & - & - \\
BUC \cite{BUC} & AAAI'19 & 38.3 & 66.2 & 79.6 & 84.5 & 27.5 & 47.4 & 62.6 & 68.4 & - & - & - & - \\
\rowcolor{mygray}
DBC \cite{DBC} & BMVC'19 & 41.3 & 69.1 & 83.0 & 87.8 & 30.0 & 51.5 & 64.6 & 70.1 & - & - & - & - \\
Y. Lin \cite{tip-u1} & TIP'20 & 38.0 & 73.7 & 84.0 & 87.9 & 30.6 & 56.1 & 66.7 & 71.5 & 9.9 & 31.4 & 41.4 & 45.7 \\
\rowcolor{mygray}
SSL \cite{SSL} & CVPR'20 & 37.8 & 71.7 & 83.8 & 87.4 & 28.6 & 52.5 & 63.5 & 68.9 & - & - & - & - \\
MMCL \cite{MMCL} & CVPR'20 & 45.5 & 80.3 & 89.4 & 92.3 & 40.2 & 65.2 & 75.9 & 80.0 & 11.2 & 35.4 & 44.8 & 49.8 \\
\rowcolor{mygray}
HCT \cite{HCT} & CVPR'20 & 56.4 & 80.0 & 91.6 & 95.2 & 50.7 & 69.6 & 83.4 & 87.4 & - & - & - & - \\
JVCT+$^{\dagger}$ \cite{JVCT} & ECCV'20 & 47.5 & 79.5 & 89.2 & 91.9 & 50.7 & 74.6 & 82.9 & 85.3 & 17.3 & 43.1 & 53.8 & 59.4 \\
\rowcolor{mygray}
CycAs$^{\dagger}$ \cite{CycAs} & ECCV'20 & 64.8 & 84.8 & - & - & 60.1 & 77.9 & - & - & 26.7 & 50.1 & - & -
\\
SpCL \cite{SpCL} & NeurIPS'20 & 73.1 & 88.1 & 95.1 & 97.0 & 65.3 & 81.2 & \textcolor{blue}{\textbf{90.3}} & 92.2 & 19.1 & 42.3 & 55.6 & 61.2 \\
\rowcolor{mygray}
CAP$^{\dagger}$ \cite{CAP} & AAAI'21 & \textcolor{blue}{\textbf{79.2}} & 91.4 & 96.3 & 97.7 & \textcolor{blue}{\textbf{67.3}} & 81.1 & 89.3 & 91.8 & \textcolor{red}{\textbf{36.9}} & \textcolor{red}{\textbf{67.4}} & \textcolor{red}{\textbf{78.0}} & \textcolor{red}{\textbf{81.4}} \\
F. Yang$^{\dagger}$ \cite{DSCE} & CVPR'21 & 61.7 & 83.9 & 92.3 & - & 53.8 & 73.8 & 84.2 & - & 15.5 & 35.2 & 48.3 & - \\
\rowcolor{mygray}
GCL$^{\dagger}$ \cite{GCL} & CVPR'21 & 66.8 & 87.3 & 93.5 & 95.5 & 62.8 & \textcolor{red}{\textbf{82.9}} & 87.1 & 88.5 & 21.3 & 45.7 & 58.6 & 64.5 \\
IICS$^{\dagger}$ \cite{IICS} & CVPR'21 & 72.9 & 89.5 & 95.2 & 97.0 & 64.4 & 80.0 & 89.0 & 91.6 & \textcolor{mygreen}{\textbf{26.9}} & \textcolor{mygreen}{\textbf{56.4}} & \textcolor{mygreen}{\textbf{68.8}} & \textcolor{mygreen}{\textbf{73.4}} \\
\rowcolor{mygray}
MPRD \cite{MPRD} & ICCV'21 & 51.1 & 83.0 & 91.3 & 93.6 & 43.7 & 67.4 & 78.7 & 81.8 & 14.6 & 37.7 & 51.3 & 57.1 \\
HCD \cite{HCD} & ICCV'21 & 78.1 & 91.1 & 96.4 & 97.7 & 65.6 & 79.8 & 88.6 & 91.6 & \textcolor{mygreen}{\textbf{26.9}} & 53.7 & 65.3 & 70.2 \\
\rowcolor{mygray}
HCD$^{\ast}$ \cite{HCD} & ICCV'21 & 77.7 & 90.9 & 96.6 & 97.6 & 65.2 & 79.5 & 89.1 & 91.9 & 22.1 & 46.7 & 58.9 & 65.2 \\
ICE \cite{ICE} & ICCV'21 & \textcolor{red}{\textbf{79.5}} & \textcolor{blue}{\textbf{92.0}} & \textcolor{blue}{\textbf{97.0}} & \textcolor{red}{\textbf{98.1}} & \textcolor{mygreen}{\textbf{67.2}} & \textcolor{mygreen}{\textbf{81.3}} & \textcolor{mygreen}{\textbf{90.1}} & \textcolor{blue}{\textbf{93.0}} & \textcolor{blue}{\textbf{29.8}} & \textcolor{blue}{\textbf{59.0}} & \textcolor{blue}{\textbf{71.7}} & \textcolor{blue}{\textbf{77.0}} \\
\rowcolor{mygray}
ICE$^{\ast}$ \cite{ICE} & ICCV'21 & 78.9 & \textcolor{mygreen}{\textbf{91.7}} & \textcolor{red}{\textbf{\underline{97.1}}} & \textcolor{mygreen}{\textbf{97.7}} & 66.4 & 80.3 & 89.6 & \textcolor{mygreen}{\textbf{92.9}} & 22.7 & 48.4 & 61.1 & 67.0 \\
\specialrule{0.1em}{0pt}{0pt}  
\textbf{Ours} & This paper & \textcolor{blue}{\textbf{\underline{79.2}}} & \textcolor{red}{\textbf{\underline{92.3}}} & \textcolor{mygreen}{\textbf{96.6}} & \textcolor{blue}{\textbf{\underline{97.8}}} & \textcolor{red}{\textbf{\underline{69.1}}} & \textcolor{blue}{\textbf{\underline{82.7}}} & \textcolor{red}{\textbf{\underline{91.1}}} & \textcolor{red}{\textbf{\underline{93.5}}} & \textbf{\underline{24.6}} & \textbf{\underline{56.2}} & \textbf{\underline{67.3}} & \textbf{\underline{71.5}} \\
\specialrule{0.1em}{0pt}{0pt}  
\end{tabular}
\end{center}
\end{table*}

\subsection{Ablation Study}

\subsubsection{Group size of group sampling} \label{sec:param-N}
Table~\ref{tab:parameter} reports the analysis of group size $N$ in group sampling. As discussed in Sec.~\ref{sec:group-sampling}, $N$ is the number of samples in each group, which plays an important role in affecting the optimization trend of the entire group. When $N$ is small, a small number of samples are grouped together, which does not enable the avoidance of consequences related to deteriorated over-fitting. When $N$ gradually increases, the group can better guide the overall trend of the class, which avoids the destruction of the similarity structure and maintains statistical stability to a certain extent, leading to better performance. 

Our experiments are conducted on Market-1501~\cite{Market} and DukeMTMC-reID~\cite{Duke}. For the characteristics of these two datasets, when $N\geq64$, the performance no longer significantly changes, and fluctuation remains at around about 1\%. This shows that at that time each group has enough samples to represent the overall class trend, and increasing the number of samples in the group will not increase the performance significantly. We believe that for different datasets, the selection of $N$ may be different. Because of the different number of images in the dataset, the number of samples in each class is not the same. Therefore, the value of $N$ needs to be set according to a specific dataset. Samples in each group need to be able to characterize the overall trend of the entire class, so it is also feasible to set $N$ to be large enough to pack all samples in the class into one group. 

\subsubsection{Shuffling degree}
In Sec.~\ref{sec:analysis}, experimental analysis shows that random sampling has a negative impact on the reliability of pseudo-labels and the generalization performance of the model, which indicates that it is unfavorable to shuffle samples with different labels during sampling. In this section, we show how shuffling impacts on the reliability of pseudo-labels and performance. On the basis of group sampling, we packed samples in adjacent $M$ mini-batches into a set, and then shuffle the samples in each set. When $M$ is larger, the shuffling degree is higher. Obviously, $M=1$ is equivalent to group sampling, and when all mini-batches are packed for shuffling, denoted as $M=all$, it is equivalent to random sampling. We take $M=\{1,4,16,64,all\}$, and the mAP score, number of clusters and NMI score are shown in Table~\ref{tab:degree-of-sampler-shuffle}. 
It shows that as the shuffling degree increases, the performance decreases significantly, the difference between the number of clusters and the ground truth continues to increase, and the NMI score also decreases. Therefore, we argue that shuffling is detrimental to feature learning, and the increased shuffling causes the amplification of defects.

\subsubsection{Influence of outliers} \label{sec:influence-outliers}
We investigate the impact of outliers on feature learning by comparing the performance of using only clustered samples and including un-clustered outliers, as shown in Table~\ref{tab:outliers}.
When training with only clustered samples, poor performance is obtained for both triplet and group sampling, suggesting that outliers play a crucial role during training.

Our contrastive baseline relies heavily on the features stored in the memory bank, which are utilized for generating pseudo-labels and calculating losses. The memory bank requires all entries to be continuously updated. Therefore, when outliers are discarded, their features cannot be updated, undoubtedly leading to training collapse.

\subsubsection{Outlier treatment in sampling}
\label{sec:treatment-of-outliers}
Sec.~\ref{sec:triplet-sampling} introduces the way that triplet sampling handles outliers, \ie, each outlier is treated as an independent group, denoted as Tre.~\uppercase\expandafter{\romannumeral1}.
Similarly, Sec.~\ref{sec:description-of-group-sampling} introduces a form of outlier treatment during group sampling, \ie, all outliers are put into the outlier set $\mathcal{O}$, and then treat $\mathcal{O}$ as a group, denoted as Tre.~\uppercase\expandafter{\romannumeral2}.
We adopt these two outlier treatment approaches for comparative experiments on triplet sampling and group sampling, as shown in Table~\ref{tab:treatment}. 

A significant degradation in performance occurs when the outlier treatment in group sampling is replaced with Tre.~\uppercase\expandafter{\romannumeral1}.
This treatment results in the possibility that un-clustered outliers and clustered instances may be in the same mini-batch, so clusters are vulnerable to outliers when there are few samples in the cluster. It suggests that Tre.~\uppercase\expandafter{\romannumeral1} is not suitable for group sampling. In this case, however, re-sampling of clustered instances in triplet sampling somewhat alleviates the impact of outliers, as confirmed by the performance of $K=16$ in Table~\ref{tab:treatment}.
Conversely, Tre.~\uppercase\expandafter{\romannumeral2} ensures that there are only un-clustered outliers or only clustered instances in a mini-batch, which reduces the mutual influence of them, and is more suitable for group sampling.
Moreover, higher performance cannot be achieved when the outlier treatment in triplet sampling is replaced with Tre.~\uppercase\expandafter{\romannumeral2}.
This suggests that the outlier treatment is not a limiting factor in obtaining higher performance with triplet sampling.

\subsection{Comparison with the State-of-the-Arts} \label{sec:sota}

We compare the proposed method with the state-of-the-art unsupervised person re-ID methods~\cite{LOMO,Market,OIM,BUC,DBC,SSL,MMCL,HCT,SpCL,tip-u1,CAP,GCL,ICE,MPRD,IICS,DSCE,CycAs,JVCT,HCD} on Market-1501~\cite{Market}, DukeMTMC-reID~\cite{Duke} and MSMT17~\cite{MSMT}, as shown in Table~\ref{tab:performance}.
Note that all listed are fully unsupervised person re-ID methods that do not use labeled source data.

Compared to the state-of-the-art methods, group sampling is simple yet exhibits highly competitive performance. It shows the best performance on DukeMTMC-reID with at least 1.8\% mAP higher than the others, and the second-best performance on Market-1501. It is worth stating that our method does not require the additional camera knowledge. Many recent methods have achieved performance gains by exploiting camera knowledge, which is particularly evident on MSMT17.
For example, although \cite{HCD,ICE} report camera-agnostic performance (see lines HCD and ICE), they still utilize camera-related information in the data sampling according to the authors' code.
For a fairer comparison, we remove the camera inputs from their sampling strategies to obtain the performance under purely camera-agnostic settings (see lines HCD$^{\ast}$ and ICE$^{\ast}$).
Despite the change in sampling strategy only, performance degradation is observed on all three datasets, and is particularly noticeable on MSMT17.
Such results demonstrate the superiority of our method over current techniques under purely camera-agnostic settings.

\section{Conclusion}
In this paper, we give an in-depth analysis of sampling strategies and reveal the important role in unsupervised person re-ID. 
In addition, we introduce new concepts, including deteriorated over-fitting and statistical stability, to explain the different performance obtained by different sampling strategies applied to the contrastive baseline. Random sampling is prone to training collapse due to the deteriorated over-fitting. In contrast, the grouping operation conduces to maintain the statistical stability within the classes, resulting in relatively better performance of triplet sampling.
Inspired by this, we further propose a novel group sampling strategy for unsupervised person re-ID. 
Group sampling effectively alleviates the negative impact of individual samples on statistical stability and fully exploits the potential of the contrastive baseline.
Extensive experimental results illustrate that group sampling achieves performance comparable to state-of-the-art unsupervised person re-ID methods on Market-1501, DukeMTMC-reID and MSMT17 datasets without introducing any additional parameters and computation costs. Moreover, our method exhibits superiority over current techniques under purely camera-agnostic settings.


\ifCLASSOPTIONcaptionsoff
  \newpage
\fi

{
\bibliographystyle{IEEEtran}
\bibliography{ref}
}

\end{document}